\documentclass[english]{article}
\let\Horig\H

\usepackage[preprint,nonatbib]{nips_2018_wider_nonotice}
\usepackage[T1]{fontenc}
\usepackage[latin9]{inputenc}
\usepackage{color,colortbl}
\usepackage{babel}
\usepackage{verbatim}
\usepackage{url}
\usepackage{amsmath}
\usepackage{amssymb}
\usepackage{graphicx}
\usepackage{setspace}
\usepackage{cancel}
\usepackage{hyperref}
\usepackage{changepage}

\hypersetup{linkcolor=blue,filecolor=magenta,urlcolor=cyan} 
\usepackage{courier}

\usepackage{amsfonts}

\renewcommand{\cal}[1]{\mathcal{#1}}

\renewcommand{\H}{\cal{H}}

\begin{document}

\title{Engineering Monosemanticity in Toy Models}
\author{Adam S. Jermyn, Nicholas Schiefer, Evan Hubinger}

\maketitle
\begin{abstract}
In some neural networks, individual neurons correspond to natural ``features'' in the input.
Such \emph{monosemantic} neurons are of great help in interpretability studies, as they can be cleanly understood.
In this work we report preliminary attempts to engineer monosemanticity in toy models.
We find that models can be made more monosemantic without increasing the loss by just changing which local minimum the training process finds.
More monosemantic loss minima have moderate negative biases, and we are able to use this fact to engineer highly monosemantic models.
We are able to mechanistically interpret these models, including the residual polysemantic neurons, and uncover a simple yet surprising algorithm.
Finally, we find that providing models with more neurons per layer makes the models more monosemantic, albeit at increased computational cost.
These findings point to a number of new questions and avenues for engineering monosemanticity, which we intend to study these in future work.
\end{abstract}

\section{Introduction}\label{sec:intro}

\subsection{Motivation}

In some neural networks, individual neurons correspond to natural ``features'' in the input.
Such \emph{monosemantic} neurons are of great help in interpretability studies, as they can be cleanly understood.
By contrast, some neurons are \emph{polysemantic}, meaning that they fire in response to multiple unrelated features in the input~\cite{olah2017feature,olah2020zoom}.
Polysemantic neurons are much harder to characterize because they can serve multiple distinct functions in a network.

Recently, \cite{elhage2022solu} and \cite{https://doi.org/10.48550/arxiv.2210.01892} demonstrated that architectural choices can affect monosemanticity, raising the prospect that we might be able to engineer models to be more monosemantic.
In this work we report preliminary attempts to engineer monosemanticity in toy models.

\subsection{Overview}

The simplest architecture that we could study is a one-layer model.
However, a core question we wish to answer is: how does the number of neurons (nonlinear units) affect the degree of monosemanticity?
To that end, we use a two-layer architecture (Section~\ref{sec:setup}).
The first layer is a linear transformation with a bias, followed by a nonlinearity.
The second layer is a linear transformation with no bias.
Our toy model is most similar to that of~\cite{XYZ}, with the key difference being that the extra linear layer allows us to vary the number of neurons independent of the number of features or the input dimension.

We study this two model on three tasks (Section~\ref{sec:setup}). The first, a feature decoder, performs a compressed sensing reconstruction of features that were randomly and lossily projected into a low-dimensional space. The second, a random re-projector, reconstructs one fixed random projection of features from a different fixed random projection. The third, an absolute value calculator, performs the same compressed sensing task and then returns the absolute values of the recovered features.
These tasks have the important property that we know which features are naturally useful, and so can easily measure the extent to which neurons are monosemantic or polysemantic (Section~\ref{sec:measures}).

\begin{figure}
\begin{adjustwidth}{-2.5cm}{-1.5cm}
\centering
\includegraphics[width=\linewidth]{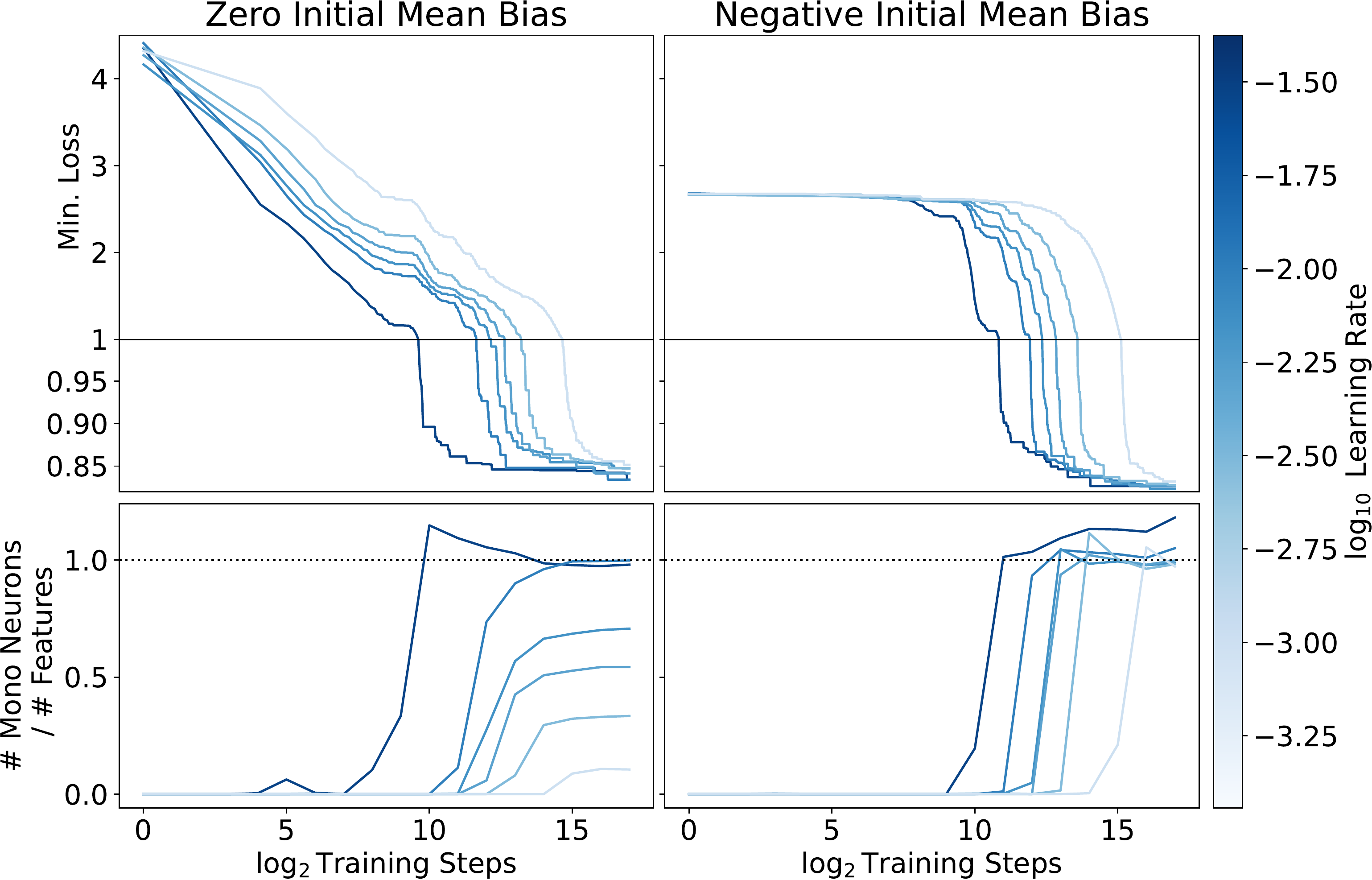}
\end{adjustwidth}
\caption{Models initialized with zero mean bias (left, batch LR1) find different local minima depending on the learning rate, with more monosemantic solutions and slightly lower loss at higher learning rates. Models initialized with a negative mean bias (right, batch LR3) all find highly monosemantic local minima, and achieve slightly better loss.}
\label{fig:hero}
\end{figure}

Our key results are (Section~\ref{sec:results}):
\begin{enumerate}
	\item When inputs are feature-sparse, models can be made more monosemantic with no degredation in performance by just changing which loss minimum the training process finds (Figure~\ref{fig:hero}, left; Section~\ref{sec:basins}).
	\item More monosemantic loss minima have moderate negative biases in all three tasks, and we are able to use this fact to engineer highly monosemantic models\footnote{Interestingly, the intervention we apply uses the path-dependence of the training process, and achieves comparable performance to directly regularizing on activation sparsity (Section~\ref{sec:reg}).} (Figure~\ref{fig:hero}, right; Section~\ref{sec:bias}).
	\item Providing models with more neurons per layer makes the models more monosemantic, albeit at increased computational cost (Figure~\ref{fig:ReLU_equal_k_training_negative_bias}; Section~\ref{sec:neurons}).
	\item These conclusions hold for a variety of activation functions and feature frequency distributions.
\end{enumerate}

In Section~\ref{sec:interp} we provide some mechanistic interpretability results for our feature decoder models in the monosemantic limit.
In particular, we found that:
\begin{enumerate}
	\item When there is a single monosemantic neuron for a feature, that neuron implements a simple algorithm of balancing feature recovery against interference (Figure~\ref{fig:amplitude_plot}, upper).
	\item When there are two monosemantic neurons for a feature, those neurons together implement an algorithm that classifies potential features as ``likely real'' or ``likely interference'', and then recovers the strength of any ``likely real'' features (Figure~\ref{fig:amplitude_plot}, lower).
	\item The remaining polysemantic neurons primarily serve to implement a linear function that adjusts the model's confidence in its output (Figure~\ref{fig:poly_linear}).
\end{enumerate}
We have not performed an equivalent analysis on the highly-polysemantic models nor on the absolute value task.
We expect that a similar kind of analysis would yield insights into the absolute value and random re-projector tasks, but are less hopeful for the highly-polysemantic models.

Finally, we have identified a number of promising avenues for future work (Section~\ref{sec:conclusions}):
\begin{enumerate}
	\item Our approach to engineering monosemanticity through bias could be made more robust by tailoring the bias weight decay on a per-neuron basis, or tying it to the rate of change of the rest of the model weights. Our sense is that there is much low-hanging fruit here.
	\item Our most polysemantic models implemenent an algorithm we do not understand.
	\item We have made naive attempts to use sparsity to reduce the cost of having more neurons per layer, but these degraded performance substantially. It is possible that deeper work in this direction will yield more workable solutions. 
\end{enumerate}

\section{Setup}\label{sec:setup}

\subsection{Features}

The training data for language models have some unusual properties:
\begin{enumerate}
	\item There are a large number of features in the data, and in particular there are more features in the data than input dimensions in the model.
	\item Each feature appears very rarely.
	\item Different features vary significantly in how frequent they are.
\end{enumerate}
Our aim in these experiments is to study toy models with a training distribution also has these properties.
Specifically, the domain we want to experiment with is one with $N$ features, which appear with average frequency $\epsilon \ll 1$, and $d$ input dimensions, such that $N \gg d \gg 1$.

To generate features with this distribution, we randomly sample $N$-dimensional vectors. Each entry corresponds to a single feature. Entry $i$ is zero with probability $1-\epsilon_i$, and otherwise is drawn uniformly over $[0,1]$.
We use two distributions of $\epsilon_i$: uniform ($\epsilon_i=\epsilon$) and power-law ($\epsilon_i \propto i^{-1.1}$, motivated by Zipf's law~\cite{Piantadosi2014}).

\subsection{Tasks}

We want to choose tasks where the features we have constructed are naturally useful.
That way there is a reason to believe that the model will encode these features so that we can e.g. check if they are associated with individual neurons.
The tasks we have chosen are a feature decoder, a random re-projector, and an absolute value calculator.
The first and third tasks take inputs of dimension $d$ and produce outputs of dimension $N$, while the second task takes inputs of dimension $d$ and produces outputs with the same dimension.

\subsubsection{Feature Decoder}

For the feature decoder task, we generate a fixed random projection matrix $P$ from $N$ dimensions down into $d$ dimensions.
We then take each feature vector, apply the projection matrix, and the task is to recover the original feature vector.
That is, if a feature vector is $\vec f$ then the task is to recover $\vec f$ given $P\cdot \vec f$, and the loss is
\begin{align}
	L = \left|\left|\vec{f} - \vec{y}\right|\right|^2,
\end{align}
where $\vec{y}$ is the model output.

\subsubsection{Re-Projector}

For the re-projector task, we generate inputs $P\cdot \vec f$ as in the feature decoder. We additionally produce a second fixed random projection matrix $Q$, again from $N$ dimensions down into $d$ dimensions.
The model is tasked with predicting $Q\cdot \vec f$ given $P \cdot \vec f$, and the loss is
\begin{align}
	L = \left|Q\cdot \vec{f} - \vec{y}\right|^2.
\end{align}

\subsubsection{Absolute Value Calculator}

For the absolute value calculator, we first sample two feature vectors $\vec{f}_1$ and $\vec{f}_2$.
The task is then to recover the element-wise absolute value $|\vec{f}_1-\vec{f}_2|$ given $P\cdot (\vec{f}_1-\vec{f}_2)$.
The loss is
\begin{align}
	L = \left|\left|\vec{|f|} - \vec{y}\right|\right|^2,
\end{align}
where $\vec{y}$ is again the model output.

\subsection{Architecture}

We study a two-layer model:
\begin{align}
	\vec{e} &= L_1 \cdot \vec{x} + \vec{b},\\
	\vec{h} &= N[\vec{e}],\\
	\vec{y} &= L_2 \cdot \vec{h}.
\end{align}
Here the input is $\vec x = P\cdot \vec{f}$.
This passes through a linear layer $L_1$ of shape $(d,k)$ and bias $\vec{b}$, a nonlinear activation function $N$ with $k$ units, and then a linear layer $L_2$ with no bias.
For the feature decoder and absolute value tasks the second linear layer $L_2$ has shape $(k,N)$, while for the re-projector task it has shape $(k,d)$.
The size $k$ of the nonlinear layer can be set independent of the input dimension $d$ and output dimension $N$, and we will use to study how $k$ affects model performance and polysemanticity.

For the most part the nonlinearity will be a ReLU, but we will also study other nonlinearities including GeLU.

\section{Measuring Monosemanticity}\label{sec:measures}

Let $F_i$ be a feature vector with feature $i$ active at unit stregth.
As a rough measure of how monosemantic a neuron is, call
\begin{align}
	r_i \equiv \frac{\max_j(h_i(F_j))}{\delta + \sum_j \max(0,h_i(F_j))},
	\label{eq:r}
\end{align}
where $h_i$ is the activation of neuron $i$, so $r_i$ is the strength of the activation of neuron $i$ to its most strongly activating feature divided by the sum of its activations over all features.
The term $\delta = 10^{-10}$ in the denominator prevents this measure from blowing up when a neuron is never active, and $\max(0,h_i)$ ignores negative values in the sum, which become relevant with some activations (e.g. GeLU) for which inactive neurons emit negative values.
When studying the absolute value task we modify equation~\eqref{eq:r} by substituting $h_i(F_j)\rightarrow h_i(F_j)+h_i(-F_j)$, as some neurons learn to be sensitive to negative inputs.

\begin{figure}
\centering
\includegraphics[width=\textwidth]{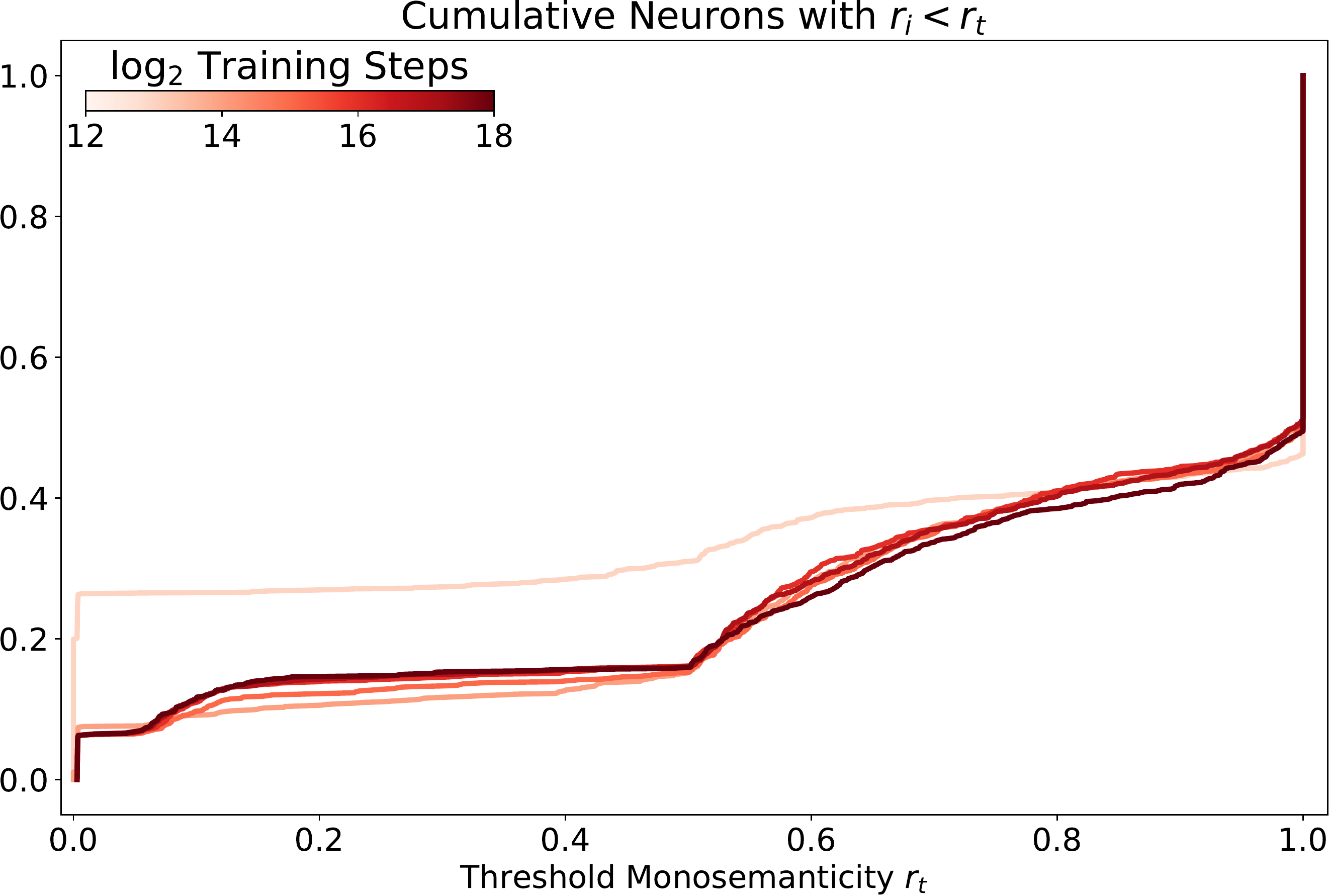}
\caption{The cumulative fraction of neurons with monosemanticity $r_i < r_t$ for varying thresholds $r_t$ for a model from batch K1 with 1024 neurons. The sharp spike at $r_t = 1$ reflects a population of precisely monosemantic neurons.}
\label{fig:measure}
\end{figure}

Figure~\ref{fig:measure} shows the cumulative density function of $r_i$ for the neurons in a model at different stages during training.
The sharp spike at $r_t = 1$ reflects a population of precisely monosemantic neurons.
In what follows we will generally call neuron $i$ monosemantic if $r_i > 0.999$.
The exact threshold is an arbitrary choice, but the measure captures an intuitive notion of monosemanticity and choosing a threshold allows us to study trends as we vary properties of the model and feature distribution.

\section{Results}\label{sec:results}

We studied our two tasks for a variety of different models.
Everywhere we used feature dimension $N=512$.
Except where we specifically state otherwise, we used embedding dimension $d=64$, ReLU activations, and feature frequency $\epsilon = 1/64$, producing an expectation of $N\epsilon = 8$ active features per sample.
We also default to studying the uniform frequency case, and call out the feature frequency distribution where we study power-law distributions.
In all cases models were trained well past the loss plateau.
Further details on the training process and the models trained are provided in Appendix~\ref{sec:training}.

For the decoder task we find that:
\begin{enumerate}
	\item \label{claim:multiple_basins1} When the nonlinear dimension $k$ exceeds the number of features $N$, these models have loss basins ranging from mostly polysemantic to mostly monosemantic (Section~\ref{sec:basins}; Figure~\ref{fig:ReLU_equal_lr_training_zero_bias}; Figure~\ref{fig:ReLU_equal_k_training_negative_bias}).
	\item \label{claim:multiple_basins2} The loss is indistinguishable between these basins (Section~\ref{sec:basins}; Figure~\ref{fig:ReLU_equal_lr_training_zero_bias}).
	\item \label{claim:monosemantic_loss_id} The monosemantic neurons are identifiable by their moderate negative bias, while the polysemantic neurons have a small positive bias (Section~\ref{sec:basins}; Figure~\ref{fig:ReLU_equal_lr_bias_zero_bias}).
	\item \label{claim:engineering_bias} By initializing models with a negative bias, we can robustly steer training towards the monosemantic basins (Figure~\ref{fig:hero}; Section~\ref{sec:bias}; Figure~\ref{fig:ReLU_equal_lr_training_negative_bias}; Figure~\ref{fig:ReLU_power_law_lr_training_negative_bias}).
	\item \label{claim:nonlinear_width} With increasing nonlinear dimension $k$, the number of monosemantic neurons increases, and models with modestly more neurons than features support assigning every feature at least one monosemantic neuron (Section~\ref{sec:neurons}; Figure~\ref{fig:ReLU_equal_k_training_negative_bias}).
	\item \label{claim:sparse_neurons} Naively introducing model sparsity at training time results in a significantly larger loss, so naive sparsity is not enough to make models with $k \gg d$ competitive.
	\item \label{claim:poly_count} Even very monosemantic models still contain a small band of $\approx d$ highly polysemantic neurons (Figure~\ref{fig:ReLU_equal_lr_sfa_negative_bias}). These implement a low-rank approximation of the identity, which we explain in Section~\ref{sec:poly}.
	\item \label{claim:sparsity} The above conclusions hold at feature frequencies up to $\epsilon = d/N$, at which point there are more features in the typical sample than there are embedding dimensions, forcing higher polysemanticity (Section~\ref{sec:feature_sparsity}; Figure~\ref{fig:eps_sweep_mono}).
\end{enumerate}
We have evidence for these results for both uniform and power-law feature distributions, as well as both ReLU and GeLU activations, though a slightly different protocol is needed to reach the monosemantic basins with GeLU activations.
We have also done experiments with a SoLU activation, but the behavior was sufficiently different that we stopped and do not have clear results for that activation function.

The same claims mostly hold for the absolute value task.
There are a few key differences, however:
\begin{enumerate}
	\item Models of different monosemanticity achieve different loss, with lower losses for more monosemantic models.
	\item Setting a negative initial bias is insufficient to robustly produce mostly-monosemantic models, and additionally either a high learning rate or low weight decay rate is needed.
\end{enumerate}
Using less weight decay produced more monosemantic models, and restores the efficacy of setting a negative initial bias.
We believe that this is because the absolute value task is more complicated to learn than the feature decoder, and so needs to spend more training steps in the relevant range of negative bias to fall into a monosemantic basin.

In the rest of this section we walk through the evidence for these claims.

\subsection{Task: Feature Decoder}\label{sec:decoder}

\subsubsection{Monosemantic and Polysemantic Basins}\label{sec:basins}

\begin{figure}
\centering
\includegraphics[width=\textwidth]{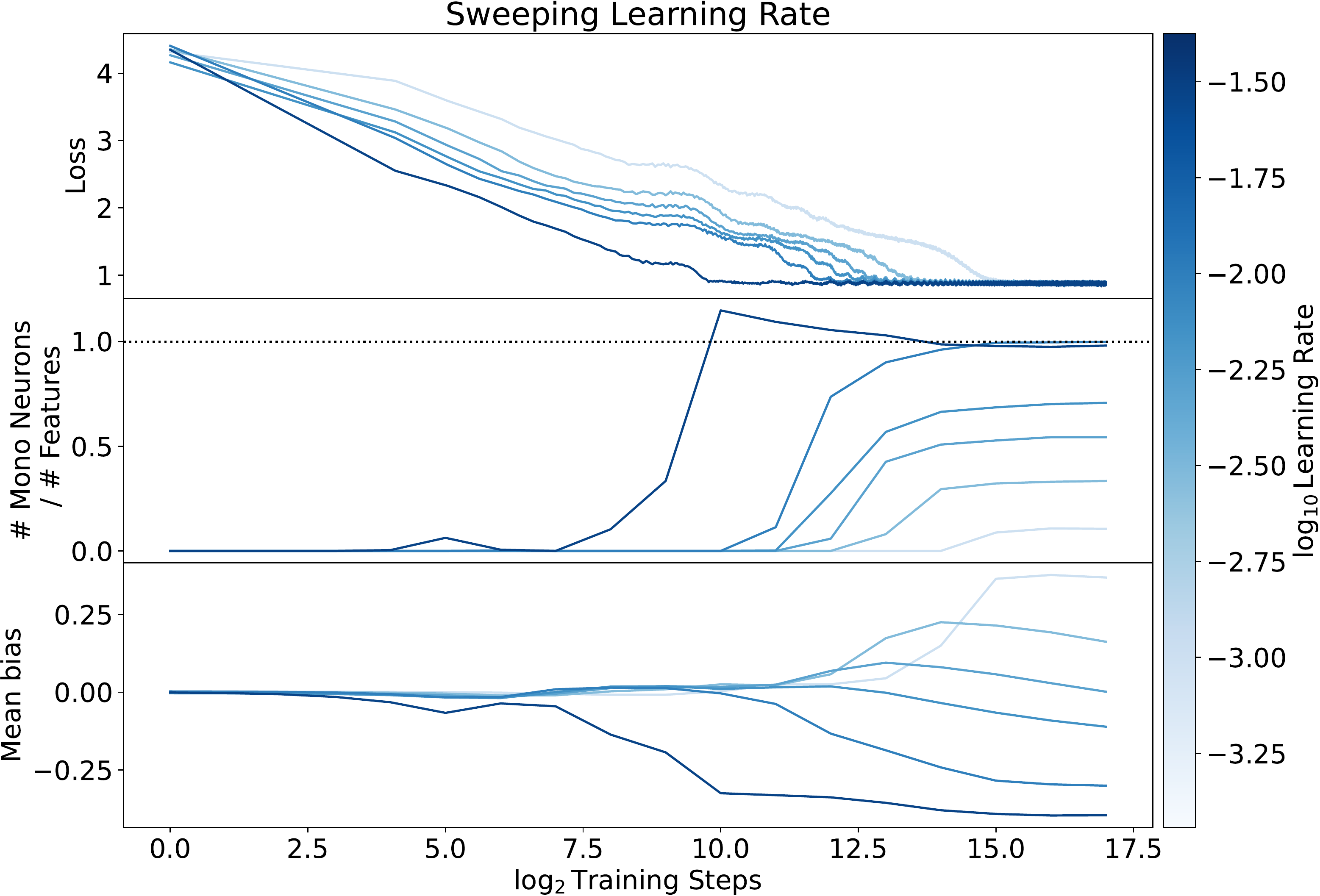}
\caption{Training traces are shown for models with different learning rates from batch LR1. Remarkably, the final models have the same loss but very different numbers of monosemantic neurons. Note that because the training samples are randomly generated anew for each batch the expected training and test losses are the same.}
\label{fig:ReLU_equal_lr_training_zero_bias}
\end{figure}

Figure~\ref{fig:ReLU_equal_lr_training_zero_bias} shows training traces for models of different learning rates with uniform feature frequencies and nonlinear dimension $k=1024$. The models all achieve the same loss (upper panel).
Because the training samples are randomly generated anew for each batch the expected training and test losses are the same.
Despite very similar performance, different models exhibit very different numbers of monosemantic neurons (middle panel) and very different average neuron bias (lower panel).
In particular, higher learning rates find loss minima with more monosemantic neurons and more negative neuron biases.

\begin{figure}
\begin{adjustwidth}{-2.5cm}{-1.5cm}
\centering
\includegraphics[width=\linewidth]{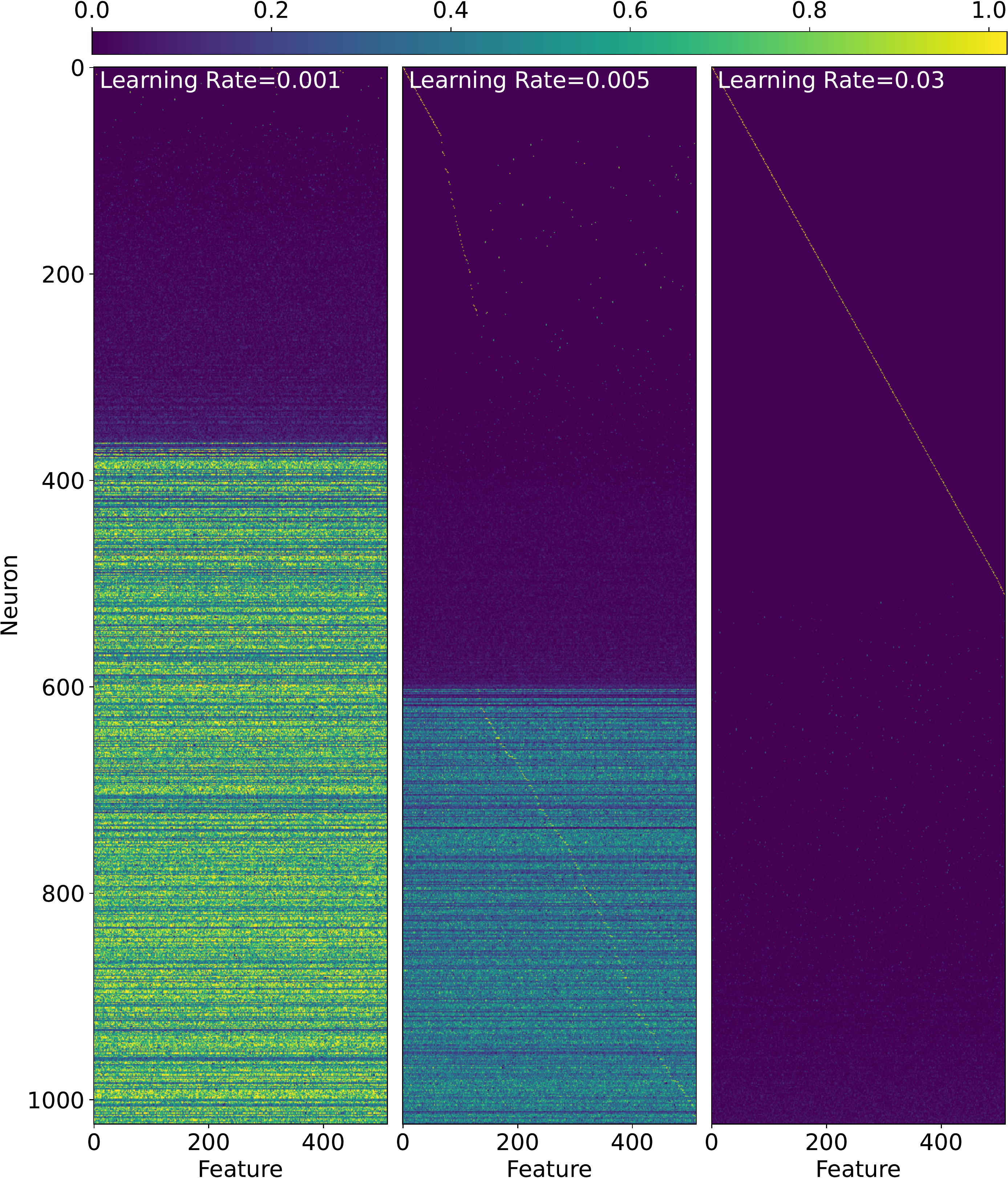}
\end{adjustwidth}
\caption{The neuron activations are shown for all single-feature inputs of unit strength. Neurons are sorted by monosemanticity, with the upper-most being the most monosemantic (equation~\ref{eq:r}). Features are sorted by the neuron they activate most-strongly. The different panels show this for different final models from batch LR1 with different learning rates. These three models implement very different algorithms but achieve the same loss.}
\label{fig:ReLU_equal_lr_sfa_zero_bias}
\end{figure}

To confirm that our measure of monosemanticity actually captures what we want, Figure~\ref{fig:ReLU_equal_lr_sfa_zero_bias} shows the activations of every neuron for all possible inputs with just one feature active at unit strength.
Neurons are sorted by monosemanticity per equation~\eqref{eq:r}, with the upper-most being the most monosemantic, and features are sorted by the neuron they activate most strongly.
The panels correspond to final models from Figure~\ref{fig:ReLU_equal_lr_training_zero_bias} trained with three different learning rates.
We see that the higher up neurons are indeed more monosemantic, with many firing for just a single feature, while those lower down can be quite polysemantic, firing on a large fraction of features.
These results support claims~\ref{claim:multiple_basins1} and \ref{claim:multiple_basins2}.

Beyond providing a sanity check, Figure~\ref{fig:ReLU_equal_lr_sfa_zero_bias} shows that the models are very different: ones trained with low learning rates have many polysemantic neurons, while those trained with higher learning rates show clean lines of monosemantic neurons, each firing in response to single features.
We emphasize that these models achieved the same loss despite implementing strikingly different algorithms.

Moreover, notice that some models have more monosemantic neurons than features.
For instance, consider the model with the highest learning rate in Figure~\ref{fig:ReLU_equal_lr_training_zero_bias}.
For 503/512 features the model has assigned a single monosemantic neuron, while for 5 features it has assigned two monosemantic neurons.
The remaining 4 features were assigned mostly-monosemantic neurons with $0.9 < r_i < 0.999$.
This provides preliminary support for claim~\ref{claim:nonlinear_width}: for models with sufficiently many neurons it is possible to assign at least one monosemantic neuron to each feature with no degredation in loss.

Note that the features with multiple neurons are not encoded redundantly.
Rather, an extra neuron allows the model to better account for its confidence in the feature being truly present (rather than faked by interference).
We discuss this in more detail in Section~\ref{sec:interp}.

Strikingly, most polysemantic neurons fire for most features, meaning that these neurons are not engaged in e.g. fighting interference between individual pairs of features.
We have analyzed these firing patterns with a variety of techniques including seriation, cosine similarity, neuron clamping, and studying how the pattern changes when multiple features are present, but so far we have not been able to understand the underlying algorithm these neurons implement except in the most-monosemantic models, which we explain in Section~\ref{sec:poly}.

\begin{figure}
\centering
\includegraphics[width=\textwidth]{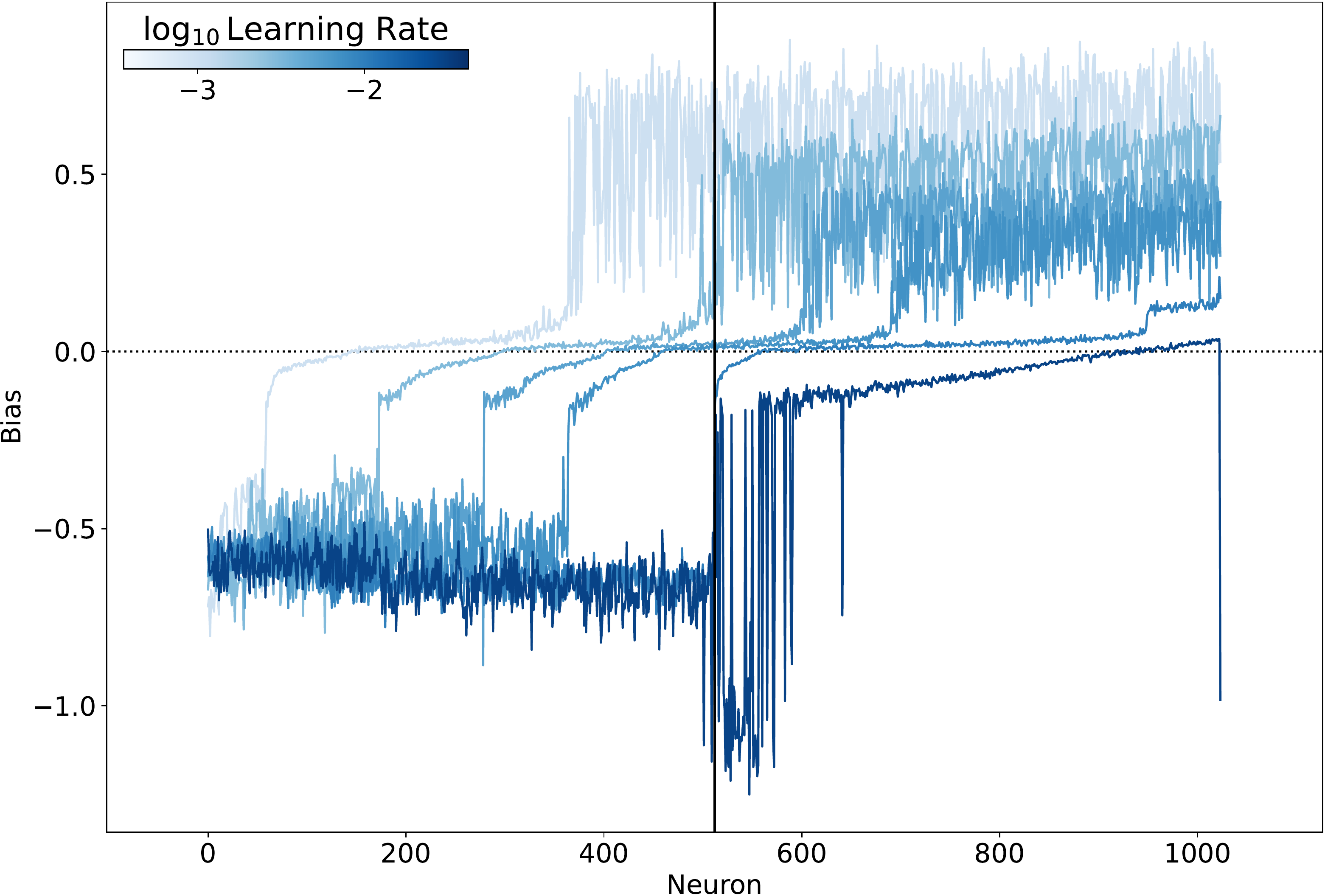}
\caption{The neuron biases are shown for the final models from batch LR1. Neurons are sorted by monosemanticity (equation~\ref{eq:r}). Curves are colored by learning rate. The most monosemantic neurons all have moderate negative biases, while the most polysemantic have moderate positive biases.}
\label{fig:ReLU_equal_lr_bias_zero_bias}
\end{figure}

It turns out that the monosemantic neurons in Figure~\ref{fig:ReLU_equal_lr_sfa_zero_bias} correspond cleanly to neurons with very negative bias.
Figure~\ref{fig:ReLU_equal_lr_bias_zero_bias} shows the bias for the final models from Figure~\ref{fig:ReLU_equal_lr_training_zero_bias}.
As before neurons are sorted by monosemanticity per equation~\eqref{eq:r} and features by the neuron they activate most strongly.
We see that the monosemantic neurons (left) all have moderate negative biases, while the polsemantic neurons (right) all have moderate positive biases.
This supports claim~\ref{claim:monosemantic_loss_id} and accords with the finding of~\cite{XYZ} that in their very similar toy model monosemantic neurons have negative bias, likely to denoise interference from unrelated features.

In between the monosemantic and polysemantic neurons is a transition region of small-magnitude biases.
The neurons in this region also have small weights, and so contribute little to the output.
We do not yet understand the purpose of these neurons.

\begin{figure}
\centering
\includegraphics[width=\textwidth]{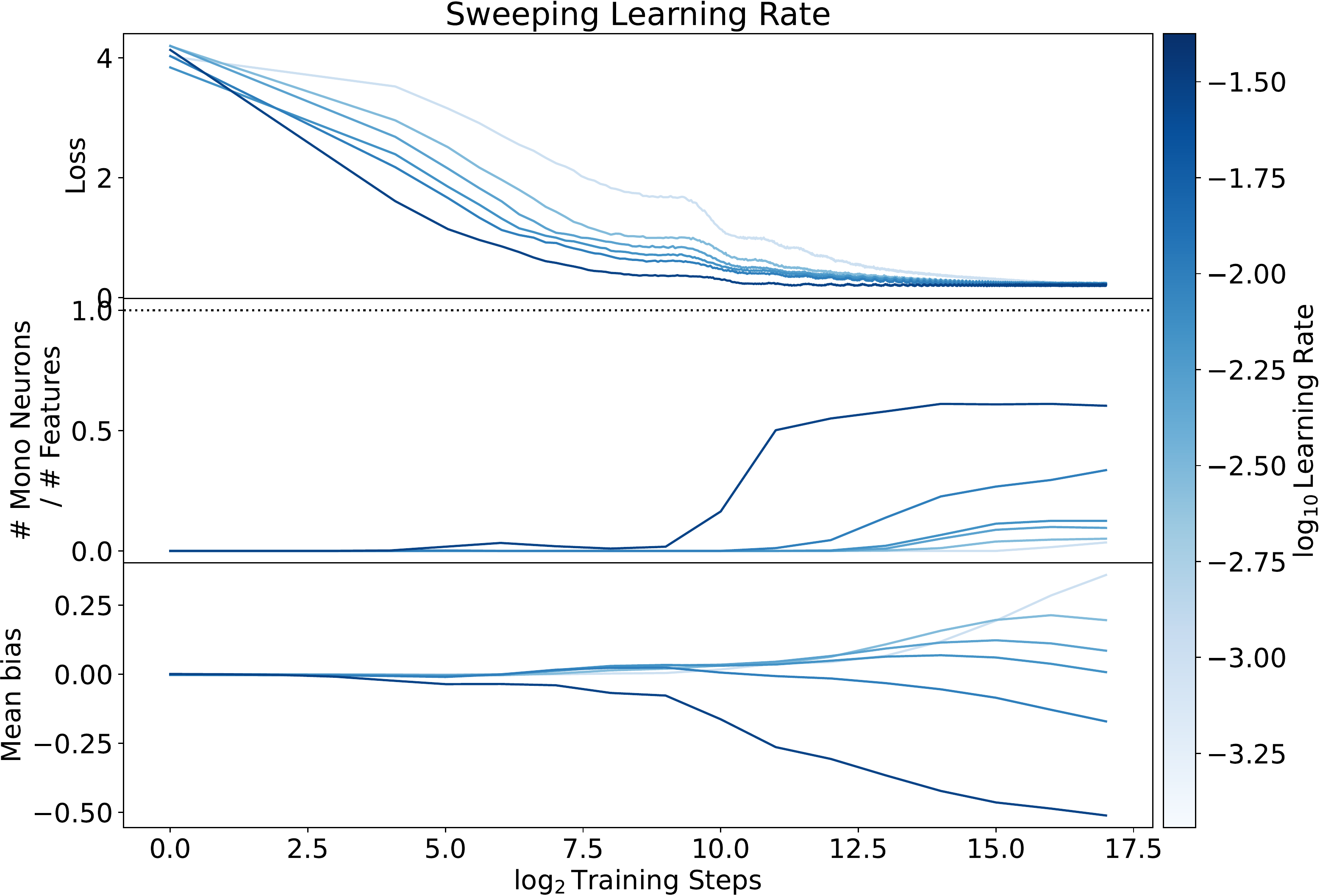}
\caption{Training traces are shown for models with different learning rates from batch LR2. Remarkably, the final models have the same loss but very different numbers of monosemantic neurons. We also see that the number of monosemantic neurons continues to rise well after the loss plateaus. This is because there are low-frequency features which make a small difference to the loss and take a long time to learn: as these are learned, the model adds monosemantic neurons but the loss does not change much. Note that because the training samples are randomly generated anew for each batch the expected training and test losses are the same.}
\label{fig:ReLU_power_law_lr_training_zero_bias}
\end{figure}

To confirm that these conclusions are not sensitive to the feature frequency distribution, we additionally trained models with power-law frequencies.
The training traces for these models are shown in Figure~\ref{fig:ReLU_power_law_lr_training_zero_bias}.
Once more we see that the most monosemantic models are those with the highest learning rates, and that these have the most negative mean biases.

Note that while we have seen various equal-loss solutions with different degrees of monosemanticity, we do not think that there is an equal-loss path in weight-space between these models.
If there were we would expect to see more evolution of monosemanticity after the loss plateau is reached, following whatever inductive biases stochastic gradient descent encodes.
Thus we are inclined to (weakly) believe that these different models reflect different basins in the loss landscape.

\subsubsection{Negative Bias and Training Protocol}\label{sec:bias}

The fact that we see negative biases in monosemantic neurons and not in polysemantic ones suggests an approach to engineering monosemanticity: initialize the biases to negative values.
We tested this by offsetting the initial small, random biases by $-1$ so that the initial mean bias is approximately $-1$.
The scale $-1$ is not arbitrary here.
The input features have uniformly distributed magnitudes in $[0,1]$ and the matrix weights are initialized to random numbers of magnitude $1/\sqrt{d}$ and random sign, so the typical input to the ReLU has scale $\sim \epsilon^{1/2}$.
This is guaranteed to be at most unity, and is quite a bit smaller for our fiducial choice of $\epsilon = 1/64$, so a bias of $-1$ is comfortably greater in magnitude than the typical activation at the start of training.

At the start of training, then, no neurons activate.
This means that the gradient of the loss vanishes and so training stalls.
To prevent this we use a small amount of weight decay on just the bias.
The weight decay gradually lowers the negative bias until neurons begin to activate, at which point there are non-zero gradients and training takes over.
Thus the weight decay allows us to start with a more negative mean bias than necessary while still ensuring that the biases eventually become small enough for neurons to activate.
Finally, to ensure that the final models are minima of the loss with no weight decay we deactivate weight decay during the final 50\% of training steps.

\begin{figure}
\centering
\includegraphics[width=\textwidth]{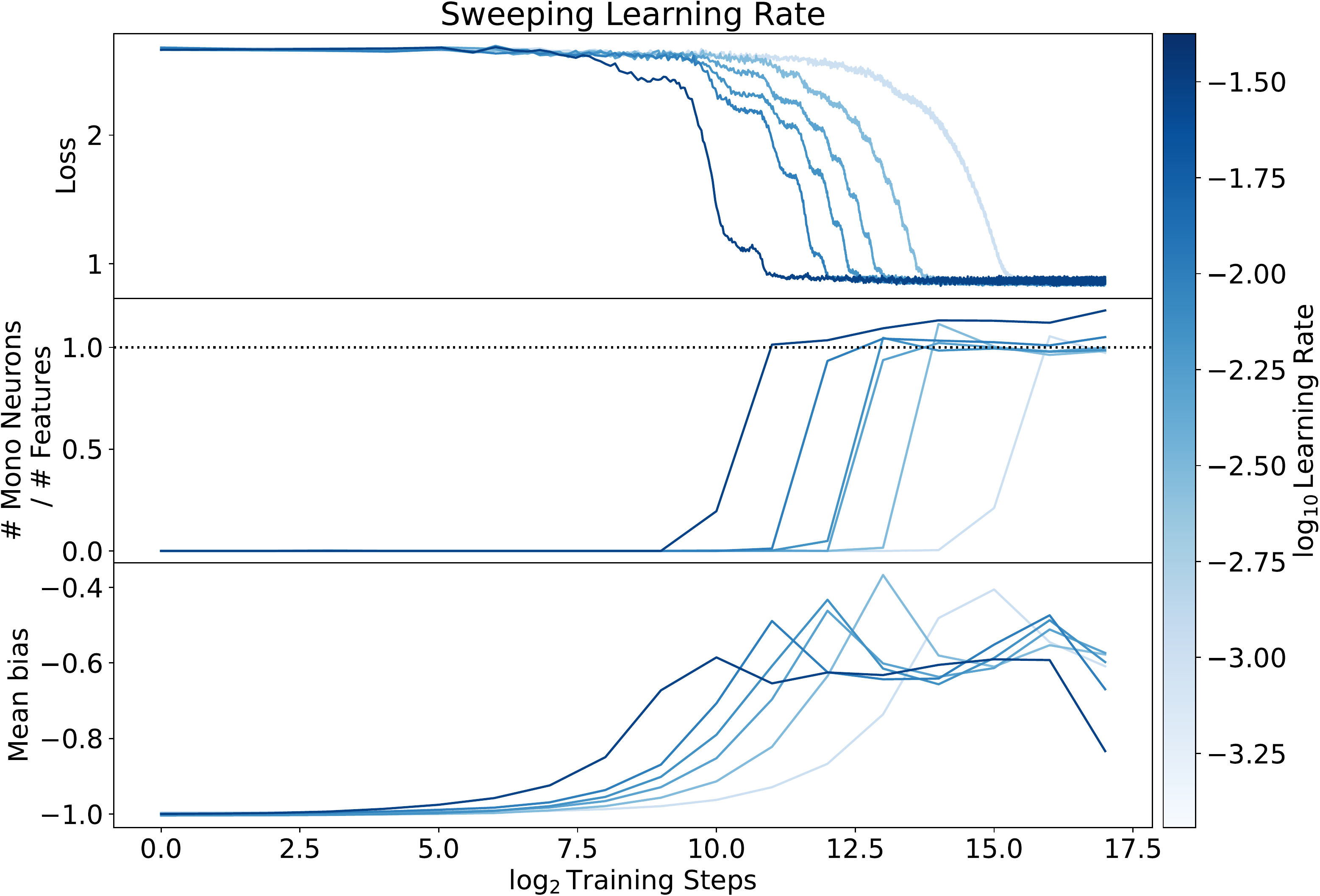}
\caption{Training traces are shown for models with different learning rates from batch LR3. Using a negative initial mean bias and weight decay allows us to ensure that models of different learning rates all find highly-monosemantic solutions.}
\label{fig:ReLU_equal_lr_training_negative_bias}
\end{figure}

Figure~\ref{fig:ReLU_equal_lr_training_negative_bias} shows that this approach indeed causes models to find very monosemantic basins even at very low learning rates.
The loss is comparable loss to models trained without this intervention, and these basins have the expected negative mean biases.
We have tried a variety of alternatives to our weight decay scheme, including approaches that adjust the bias depending on the fraction of the time that each neuron activates and approaches involving the gradients with respect to each bias.
None of the variations we have tried outperformed our simple weight decay schedule and all were more complicated, so here we only present results with weight decay as described above.

\begin{figure}
\centering
\includegraphics[width=0.85\textwidth]{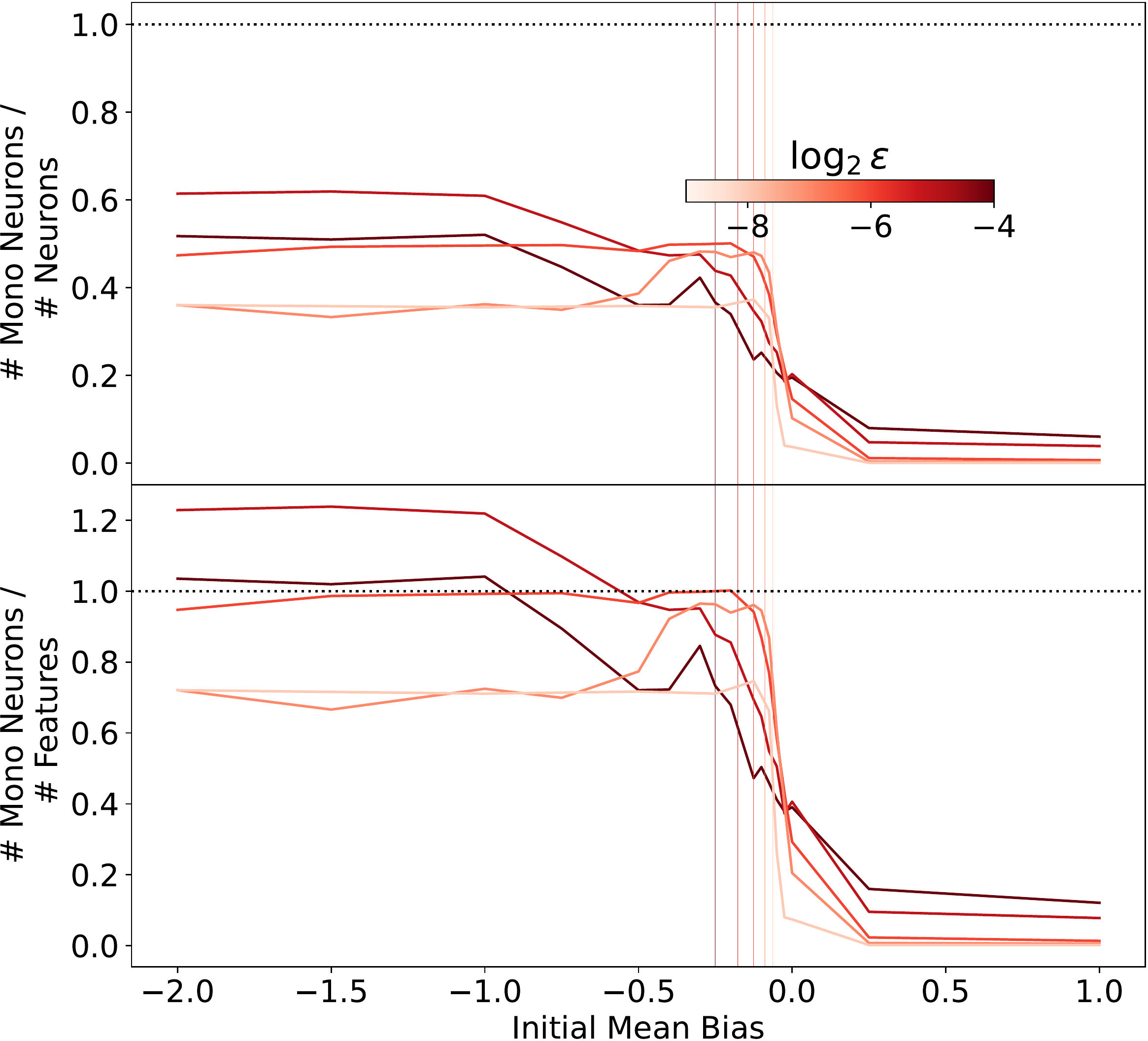}
\caption{The fraction of neurons which are monosemantic (upper) and the ratio of monosemantic neurons to input features (lower) are shown as a function of the initial mean bias and feature density $\epsilon$ for models from batches B1-B5. The dashed black lines show where the number of monosemantic neurons equals the number of neuronss (upper) and the number of features (lower). The vertical lines shows the scale $-\epsilon^{1/2}$ corresponding to each curve. Both occur near where the monosemantic fraction stars dropping, indicating that the initial mean bias becomes less effective at producing monosemantic solutions once it is more positive than $\approx -\epsilon^{1/2}$. }
\label{fig:ReLU_equal_bias_sweep_mono}
\end{figure}

\begin{figure}
\centering
\includegraphics[width=0.85\textwidth]{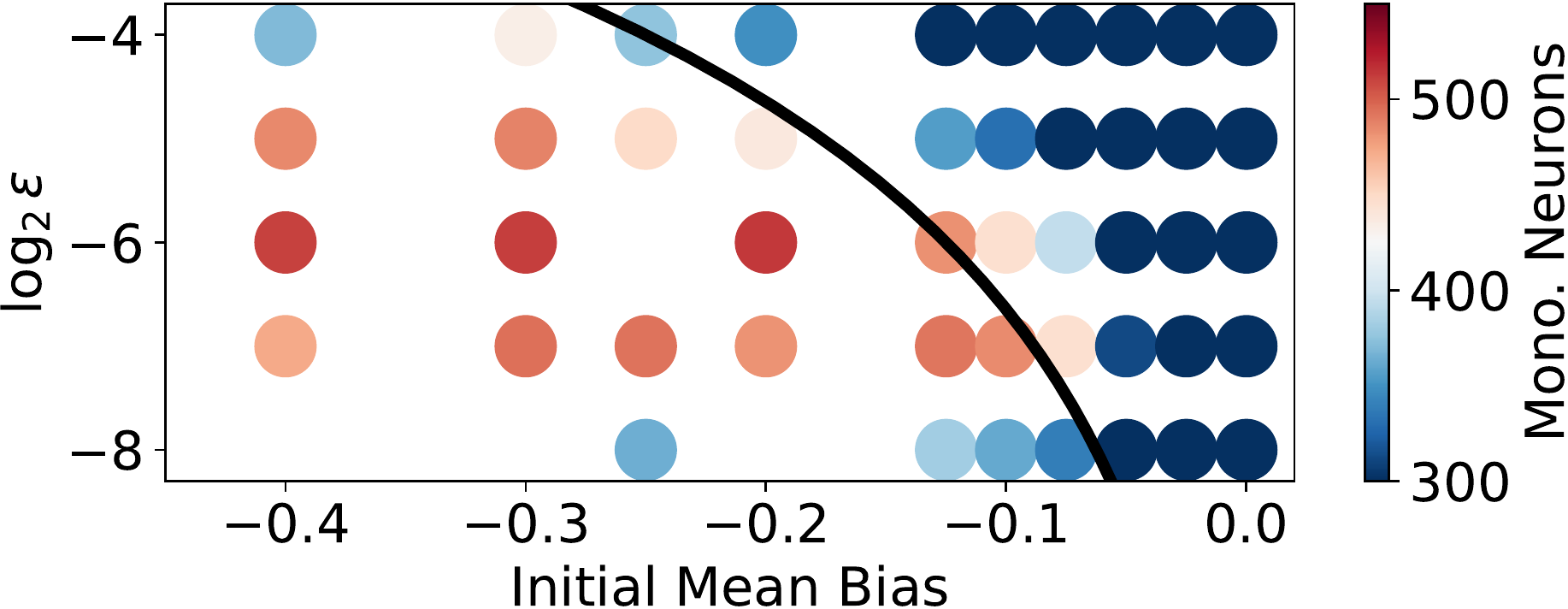}
\caption{The number of monosemantic neurons are shown as a function of the initial mean bias and feature density $\epsilon$ for models from batches B1-B5. The sharp dropoff in monosemanticity corresponds roughly to crossing the bias scale $-\epsilon^{1/2}$ (black curve), indicating that the initial mean bias becomes less effective at producing monosemantic solutions once it is more positive than $\approx -\epsilon^{1/2}$. }
\label{fig:ReLU_equal_bias_sweep_color}
\end{figure}

Figure~\ref{fig:ReLU_equal_bias_sweep_mono} provides evidence that with this weight decay approach the precise initial bias does not matter for ReLU activations so long as it is greater than the scale $\sim \epsilon^{1/2}$.
There we have plotted the fraction of neurons which are monosemantic (upper) and the ratio of monosemantic neurons to input features (lower) for models of varying initial bias.
The vertical dashed lines show the scale $-\epsilon^{1/2}$.
For initial mean biases less than $\approx -\epsilon^{1/2}$ we see that 30-50\% of the neurons are monosemantic, with 0.6-1 monosemantic neurons per feature.
The number of monosemantic neurons drops off rapidly for greater (more-positive) biases.

Figure~\ref{fig:ReLU_equal_bias_sweep_color} provides an alternate view of the same data.
There individual runs are shown as points in the space of initial mean bias and logarithmic feature frequency.
The black line shows the bias scale $-\epsilon^{1/2}$, and the shapr dropoff in monosemanticity corresponds roughly to crossing this curve.

\begin{figure}
\begin{adjustwidth}{-2.5cm}{-1.5cm}
\centering
\includegraphics[width=\linewidth]{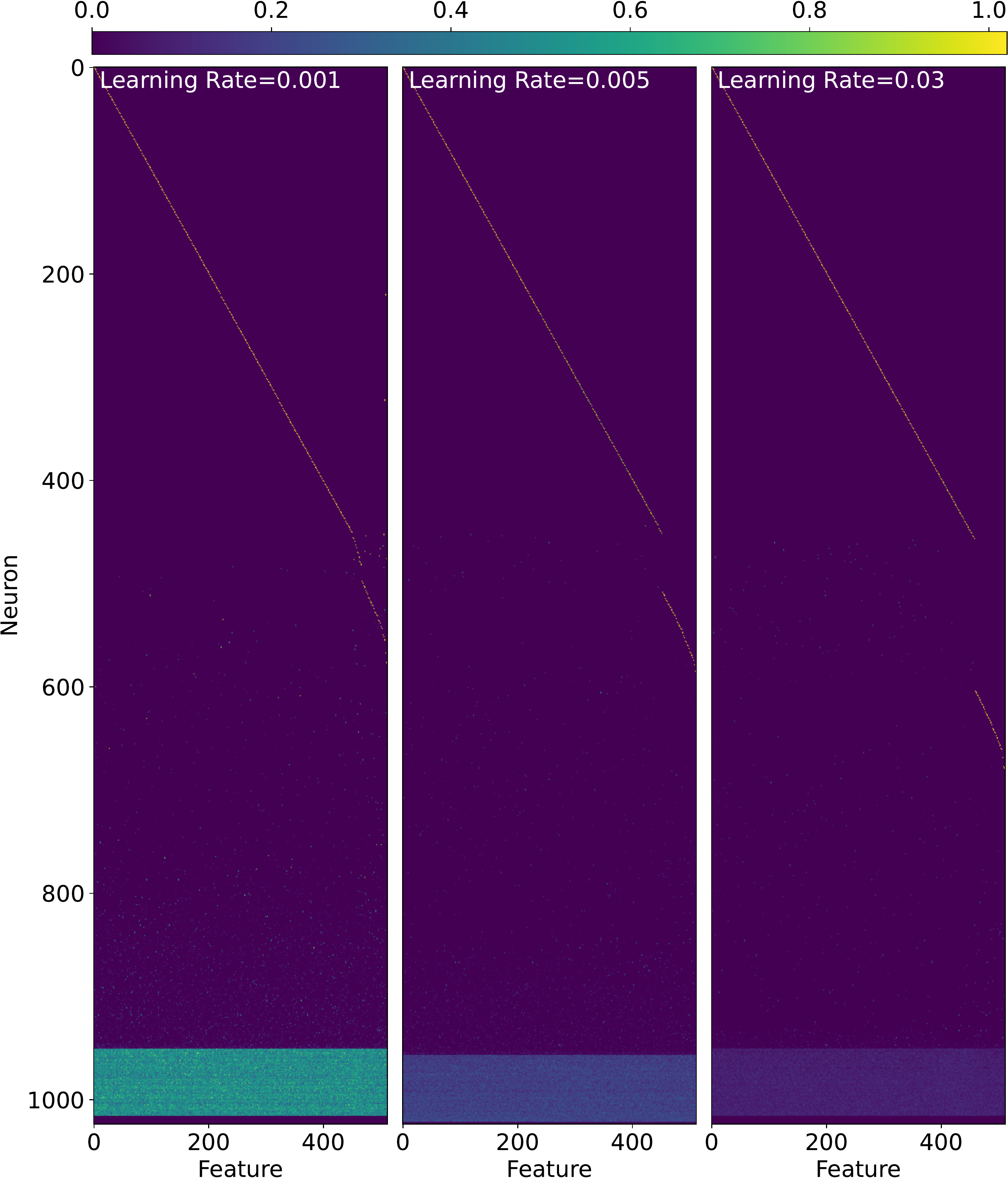}
\end{adjustwidth}
\caption{The neuron activations are shown for all single-feature inputs of unit strength for models with differen learning rates from batch LR3. Neurons are sorted by monosemanticity, with the upper-most being the most monosemantic (equation~\ref{eq:r}).  Features are sorted by the neuron they activate most-strongly. The different panels show this for different final models from Figure~\ref{fig:ReLU_equal_lr_training_negative_bias} with different learning rates. Contrasting with Figure~\ref{fig:ReLU_equal_lr_sfa_zero_bias}, we see that a negative initial bias results in highly monosemantic solutions across the board. There is still a cluster of polysemantic neurons, whose function is to produce a low-rank approximation of the identity (Section~\ref{sec:poly}).}
\label{fig:ReLU_equal_lr_sfa_negative_bias}
\end{figure}

Turning back to the models initialized with a mean bias of $-1$, we can confirm that these are indeed mostly monosemantic by plotting the single-feature activations (Figure~\ref{fig:ReLU_equal_lr_sfa_negative_bias}).
We see that the models show similar numbers of monosemantic neurons regardless of learning rate, and they far fewer polysemantic neurons than before.

\begin{figure}
\centering
\includegraphics[width=\textwidth]{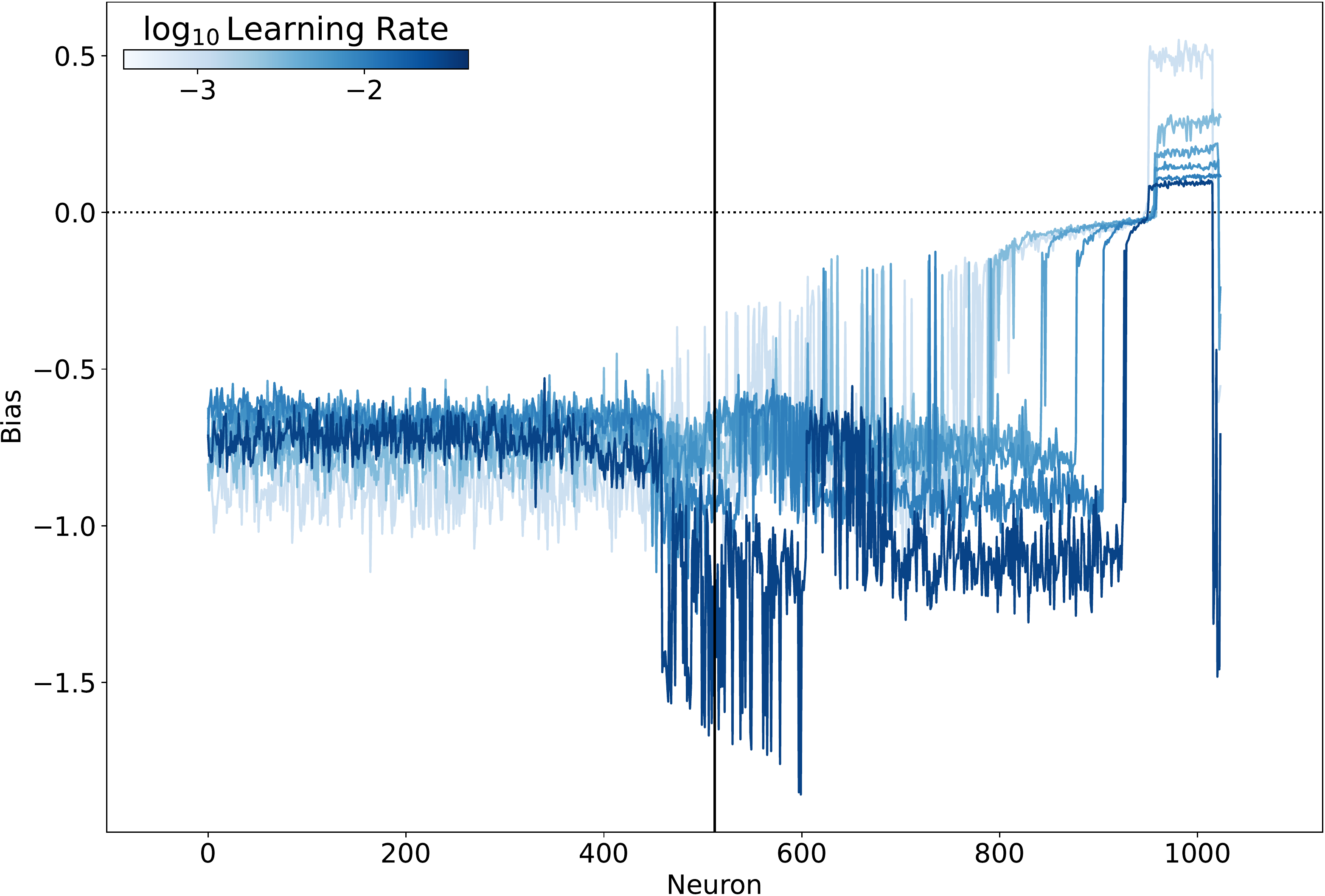}
\caption{The neuron biases are shown for the final models from batch LR3. Neurons are sorted by monosemanticity (equation~\ref{eq:r}). Curves are colored by learning rate. We see that the polysemantic neurons from Figure~\ref{fig:ReLU_equal_lr_sfa_negative_bias} correspond to positive-bias neurons, and that the monosemantic neurons correspond to those with negative biases. Remarkably, the width of the polysemantic band of neurons is uniform across models, exactly 65 neurons wide in each case, which is one more than the embedding dimension. We understand why the number of these neurons is near the embedding dimension (Section~\ref{sec:poly}) but not why there should be one more.}
\label{fig:ReLU_equal_lr_bias_negative_bias}
\end{figure}

We can inspect these neurons further by examining their biases (Figure~\ref{fig:ReLU_equal_lr_bias_negative_bias}).
Each model has exactly 65 neurons with positive biases, corresponding to the polysemantic neurons we saw in Figure~\ref{fig:ReLU_equal_lr_sfa_negative_bias}.
Strangely, this is one more than the number of embedding dimensions.
This continues to hold very closely if we vary the embedding dimension (claim~\ref{claim:poly_count}).
Specifically, with embedding dimension 32 we see 33 polysemantic neurons, and with embedding dimension 128 we see 130 polysemantic neurons, using bias above $0.05$ as a proxy for polysemanticity.
We explain why these numbers are so close to the embedding dimension $d$ in Section~\ref{sec:poly}, though we are not sure why it is always slightly in excess of $d$.

In addition to the polysemantic neurons, we again see monosemantic neurons with a moderate negative bias, as well as neurons with small-magnitude biases for which the model weights are likewise small.
There is also a new class of neurons not present in our earlier models.
These have large negative biases, and we confirmed that these neurons rarely activate, even when multiple features are present.
They appear to just not be used.
These are only present in the model with the highest learning rate of 0.03, though, and so may not be a generic feature of these models.

\begin{figure}
\centering
\includegraphics[width=\textwidth]{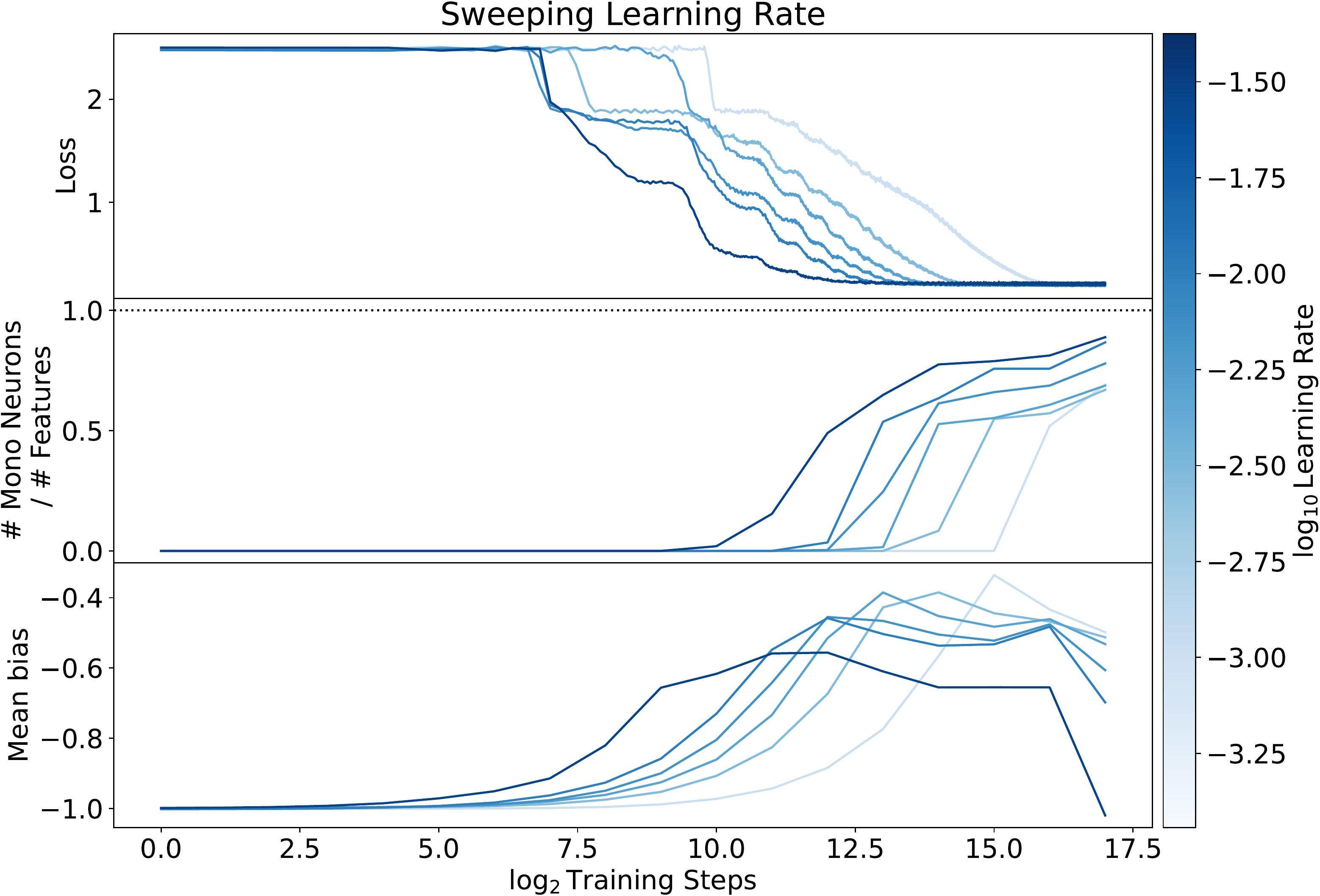}
\caption{Training traces are shown for models with different learning rates with power-law feature frequencies from batch LR4. Using a negative initial mean bias and weight decay allows us to ensure that models of different learning rates all find highly-monosemantic solutions.}
\label{fig:ReLU_power_law_lr_training_negative_bias}
\end{figure}

We confirmed that this negative bias approach works with the power-law feature frequency distribution.
Training traces produced with power-law frequencies are shown in Figure~\ref{fig:ReLU_power_law_lr_training_negative_bias}. Note that the fraction of monosemantic neurons continues going up even after the loss stops noticeably decreasing.
We suspect that this is because the most infrequent features are still being learned and so produce more monosemantic neurons, but contribute so little to the loss that they do not make an appreciable difference there.

\begin{figure}
\centering
\includegraphics[width=\textwidth]{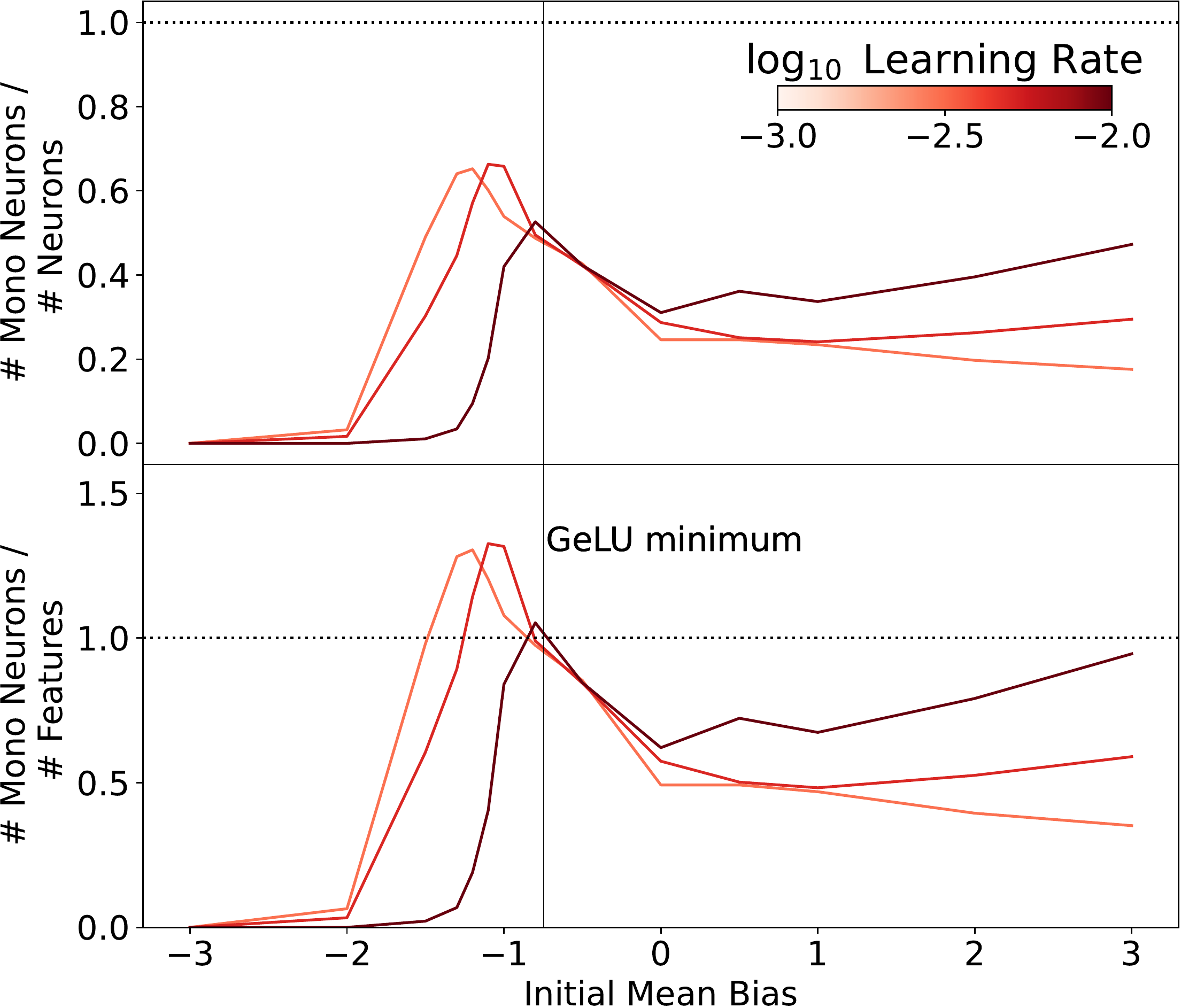}
\caption{The fraction of neurons which are monosemantic (upper) and the ratio of monosemantic neurons to input features (lower) are shown for models with GeLU activations from batch B3 as functions of the initial bias and learning rate. The dashed black lines show where the number of monosemantic neurons equals the number of neurons (upper) and the number of features (lower). The vertical line shows the location of the minimum of the GeLU. We see that the GeLU activation comes with a preferred bias scale of roughly $-1$ at which models learn highly monosemantic solutions.}
\label{fig:GeLU_equal_bias_sweep_mono}
\end{figure}

We likewise confirmed that our negative bias and weight decay approach works with GeLU activations, though there the choice of initial mean bias is more complicated.
Figure~\ref{fig:GeLU_equal_bias_sweep_mono} shows the fraction of neurons which are monosemantic (upper) and the ratio of monosemantic neurons to input features (lower) for these models as functions of the initial bias and learning rate.
The monosemanticity is sharply peaked towards an initial bias of $-1$, which lies very near the minimum in the GeLU activation function (vertical line), just slightly on the more negative side.
The peak shifts slightly with different learning rates.
We are not sure why these values are preferred, though it is not surprising that GeLU, which is not scale-invariant, introduces a preferred scale distinct from those of the initial weights and feature norms.

As a practical matter, the characteristic scale of the GeLU activation gives us a prescription for engineering monosemantic neurons, namely to initialize with an initial mean bias of approximately $-1$, and to scale the initial weights such that activations at the start of training are \emph{small} compared with that scale.
This second step is necessary to ensure that the initial bias matters and is not tiny compared with the activations at the start of training.

All told, our findings support claim~\ref{claim:engineering_bias}, namely that we can engineer monosemanticity in these models by means of an initial negative bias.

\subsubsection{Comparison with L1 Regularization}\label{sec:reg}

\begin{figure}
\centering
\includegraphics[width=\textwidth]{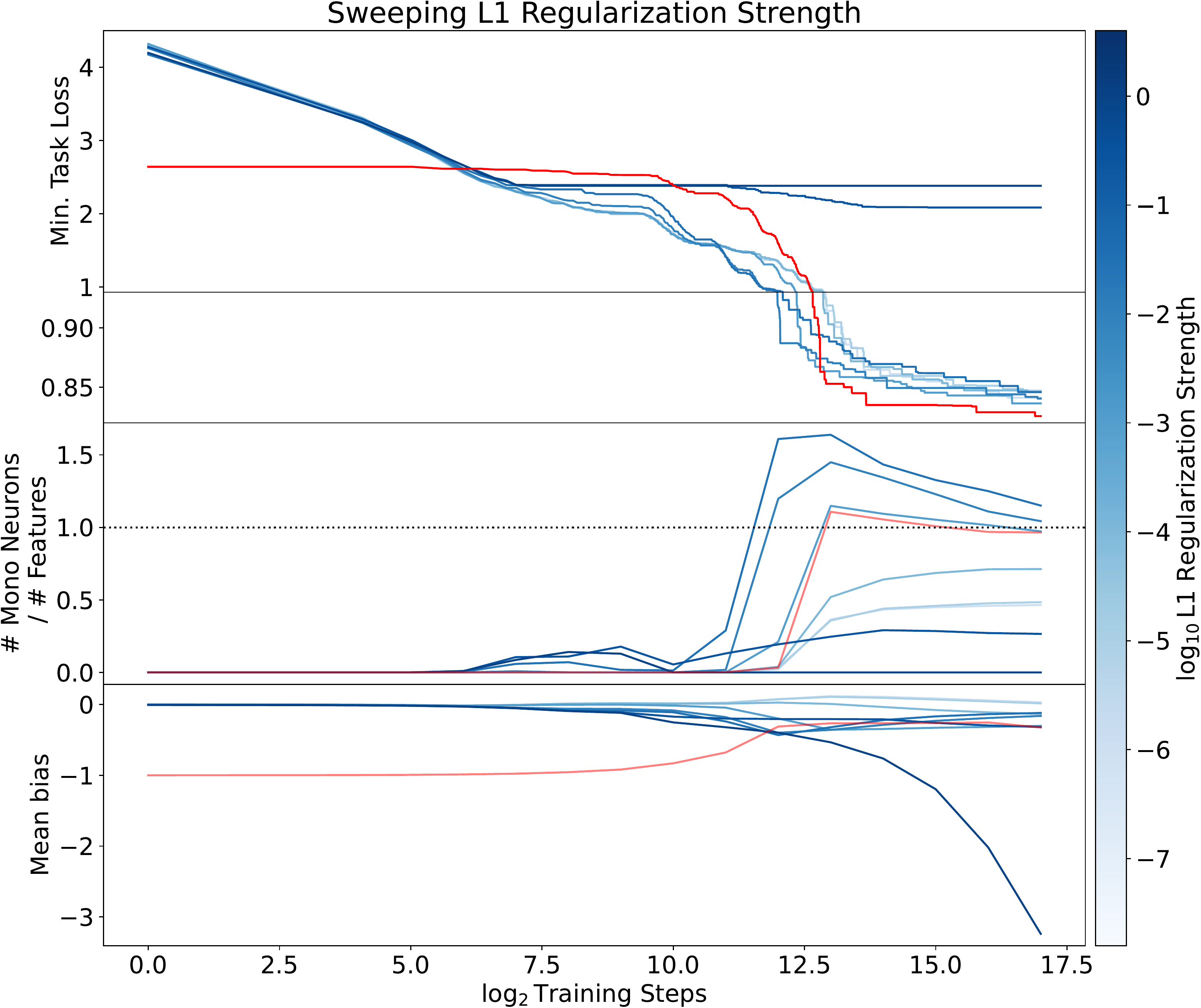}
\caption{Training traces are shown for models with different L1 regularization strengths from batch RG1, as well as a model trained with the same parameters but no regularization and a negative initial mean bias (red). The negative initial bias model achieves lower loss and comparable monosemanticity to the best model found with L1 regularization.}
\label{fig:reg_training}
\end{figure}

We compared our approach of initializing models with a negative mean bias to the more standard technique of L1 regularization on the neuron activations.
Figure~\ref{fig:reg_training} shows training traces with L1 regularization and zero initial mean bias (blue) as well as one with no regularization and a negative initial mean bias (red).
We see that the negative initial bias model achieves comparable task loss and monosemanticity to the best model found with L1 regularization.

There are different tradeoffs involved with our bias initialization scheme and L1 regularization.
With our scheme the tunable parameter is the bias decay rate, for which the optimal value seems to vary between tasks (see Section~\ref{sec:abs}).
Too high a decay rate and we see more polysemantic solutions, too low a decay rate and the training process is slow.
By contrast, the tradeoff with L1 regularization is that regularizing too strongly hurts task performance, while regularizing too weakly produces more polysemantic solutions.

We do not see these techniques as being in competition.
Rather, they are different approaches to a similar end: the bias initialization intervention leverages path-dependence in training, while the L1 regularization shifts the loss landscape to favor minima with sparser activations.
Which of these is preferable is likely context-dependent, and we do not necessarily need to choose: both interventions can be leveraged in the same model if desired.

\subsubsection{Varying Nonlinear Dimension}\label{sec:neurons}

So far we have used nonlinear dimension (i.e. number of neurons) $k=1024$ in all of our calculations.
We now study how the number of monosemantic neurons varies as we vary $k$.

\begin{figure}
\centering
\includegraphics[width=\textwidth]{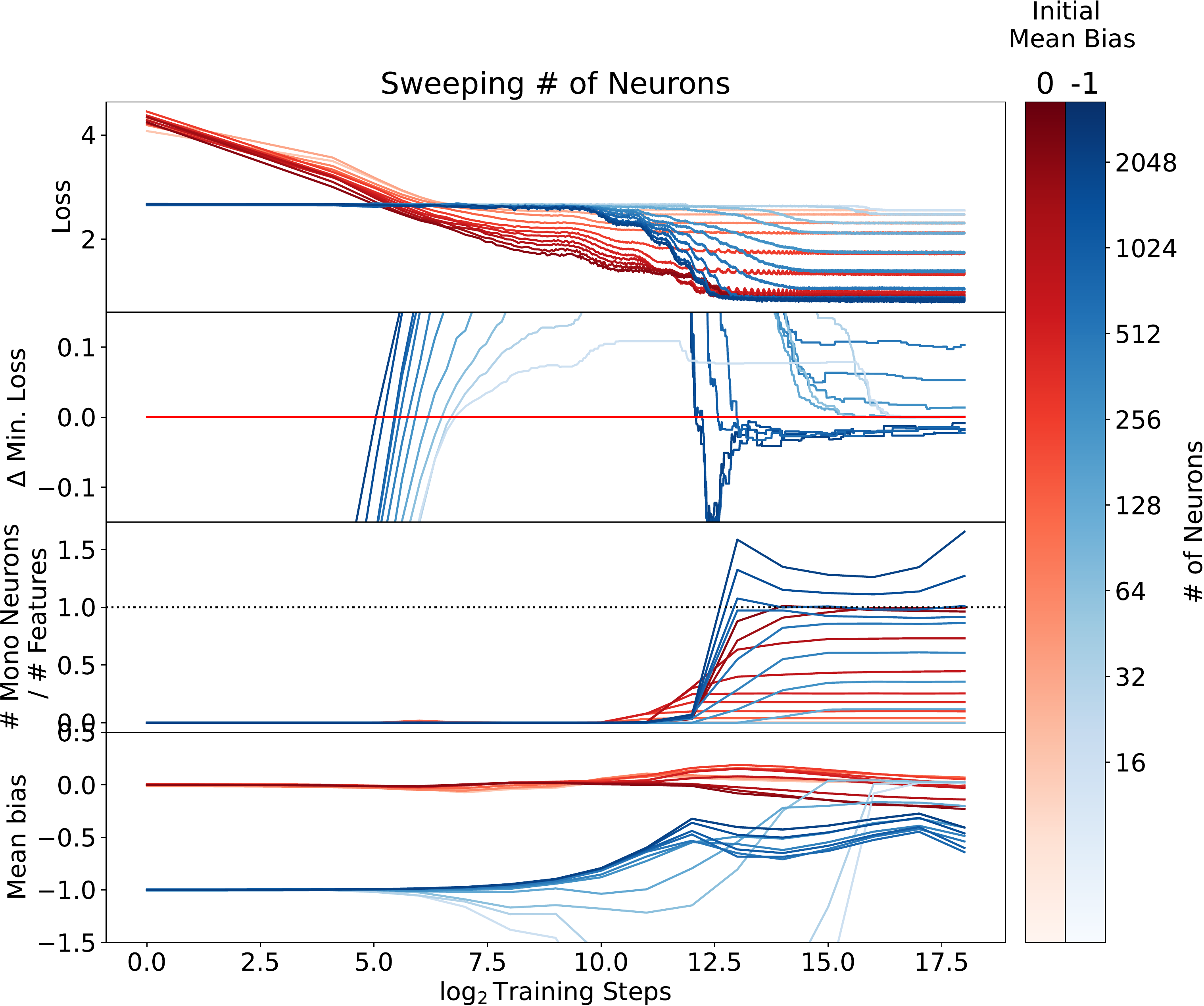}
\caption{Training traces are shown for models with different nonlinear dimensions from batch K0 with zero initial mean bias (red) and batch K1 with a negative initial mean bias (blue). Models with more neurons learn more monosemantic solutions and achieve lower loss up until the number of neurons exceeds the number of features, at which point the loss sees sharply diminishing returns with scale and additional neurons mostly just increases the number of monosemantic neurons. Models with a negative initial mean bias learn much more monosemantic representations and have very similar losses.}
\label{fig:ReLU_equal_k_training_negative_bias}
\end{figure}

\begin{figure}
\centering
\includegraphics[width=\textwidth]{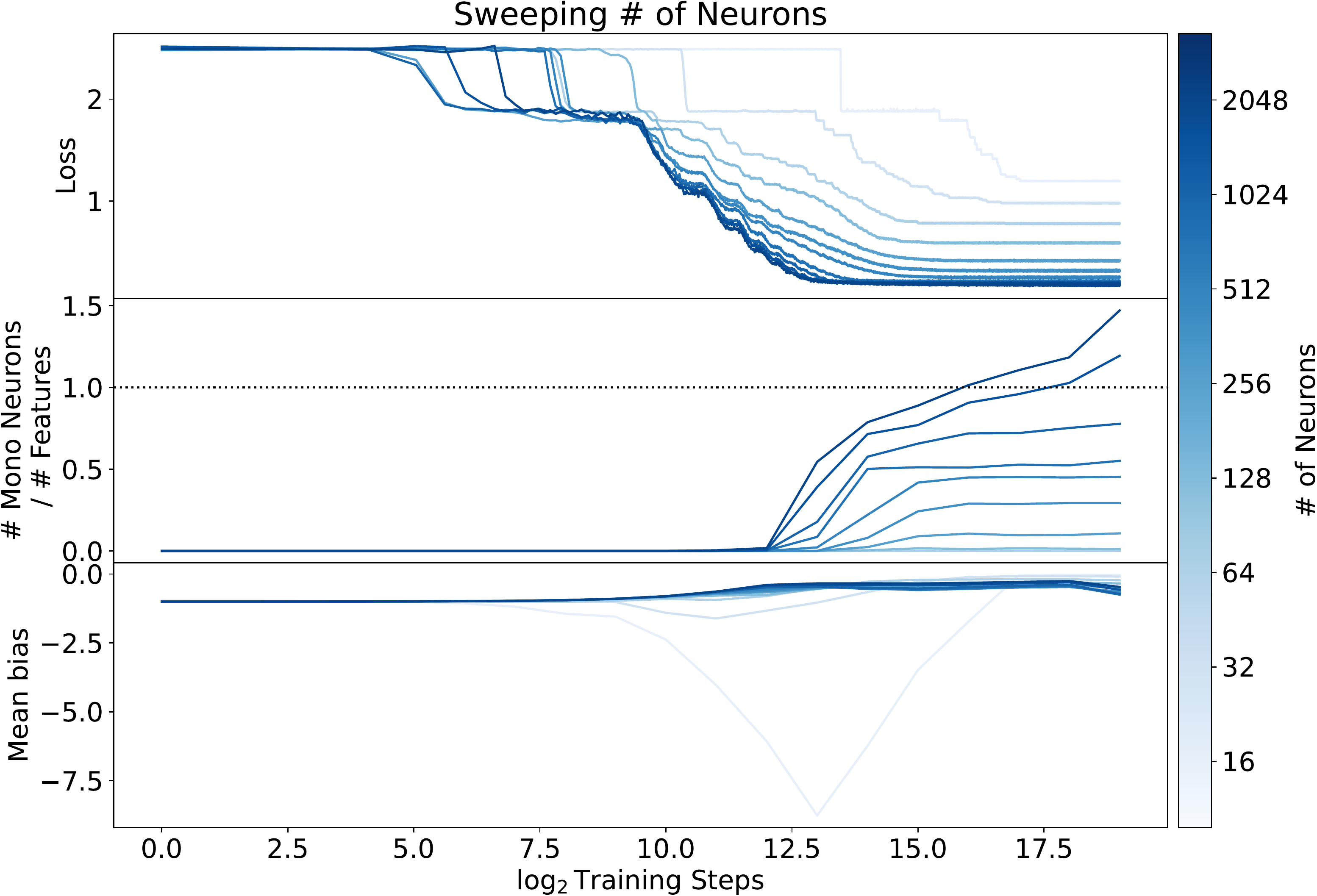}
\caption{Training traces are shown for models with different nonlinear dimensions and power-law feature frequencies from batch K2. Models with more neurons learn more monosemantic solutions and achieve lower loss up until the number of neurons exceeds the number of features, at which point the loss sees sharply diminishing returns with scale and additional neurons mostly just increases the number of monosemantic neurons. Notice that the largest models continue to gain more monosemantic neurons through additional training even after the loss plateaus: we suspect that this is because the most infrequent features are still being learned and so produce more monosemantic neurons, but contribute so little to the loss that they do not make an appreciable difference there.}
\label{fig:ReLU_pl_k_training_negative_bias}
\end{figure}

Figure~\ref{fig:ReLU_equal_k_training_negative_bias} shows training traces for models with different numbers of neurons.
Focusing just on the blue curves, which show models from batch K1 with a negative initial mean bias, we see that increasing the number of neurons in the model decreases the final loss until the number of neurons exceeds the number of features, at which point the loss is similar across models of different sizes.
Models with more neurons \emph{also} have more monosemantic neurons, and we see evidence that in the largest models the number of monosemantic neurons continues to rise even after the loss plateaus.

The trend of increasing monosemanticity with increasing training time is even more striking when features frequencies are power-law distributed (Figure~\ref{fig:ReLU_pl_k_training_negative_bias}).
We suspect that this is the same phenomenon we saw in Figure~\ref{fig:ReLU_power_law_lr_training_negative_bias}, namely that the most infrequent features are still being learned and so produce more monosemantic neurons, but contribute so little to the loss that they do not make an appreciable difference there.

Comparing the models with zero and negative initial mean biases (Figure~\ref{fig:ReLU_equal_k_training_negative_bias}, blue versus red) we see that the negative initial mean bias indeed increases the number of monosemantic neurons and usually does not come at any significant performance cost.
In particular, the middle panel shows the loss difference (blue minus red) between the two families of models.
For models with 16-128 neurons we see indistinguishable loss plateaus with different initial mean biases.
For models with 256, 384, and 512 neurons we see slightly worse performance in the more monosemantic models (negative initial mean bias).
Finally, for models with more than 512 neurons, which is also the number of features, we see that the more monosemantic models (negative initial mean bias) outperform the more polysemantic models (zero initial mean bias).

This trend of monosemantic models outperforming polysemantic ones when there are more neurons than features, and underperforming in the opposite limit, is a result of monosemantic representations being more ``natural'' for this feature-decoding task so long as there are enough neurons to assign each feature one or more monosemantic neurons.
In this (weakly-held) view, polysemanticity is the way the model copes with having too few neurons for the task at hand.

\begin{figure}
\centering
\includegraphics[width=\textwidth]{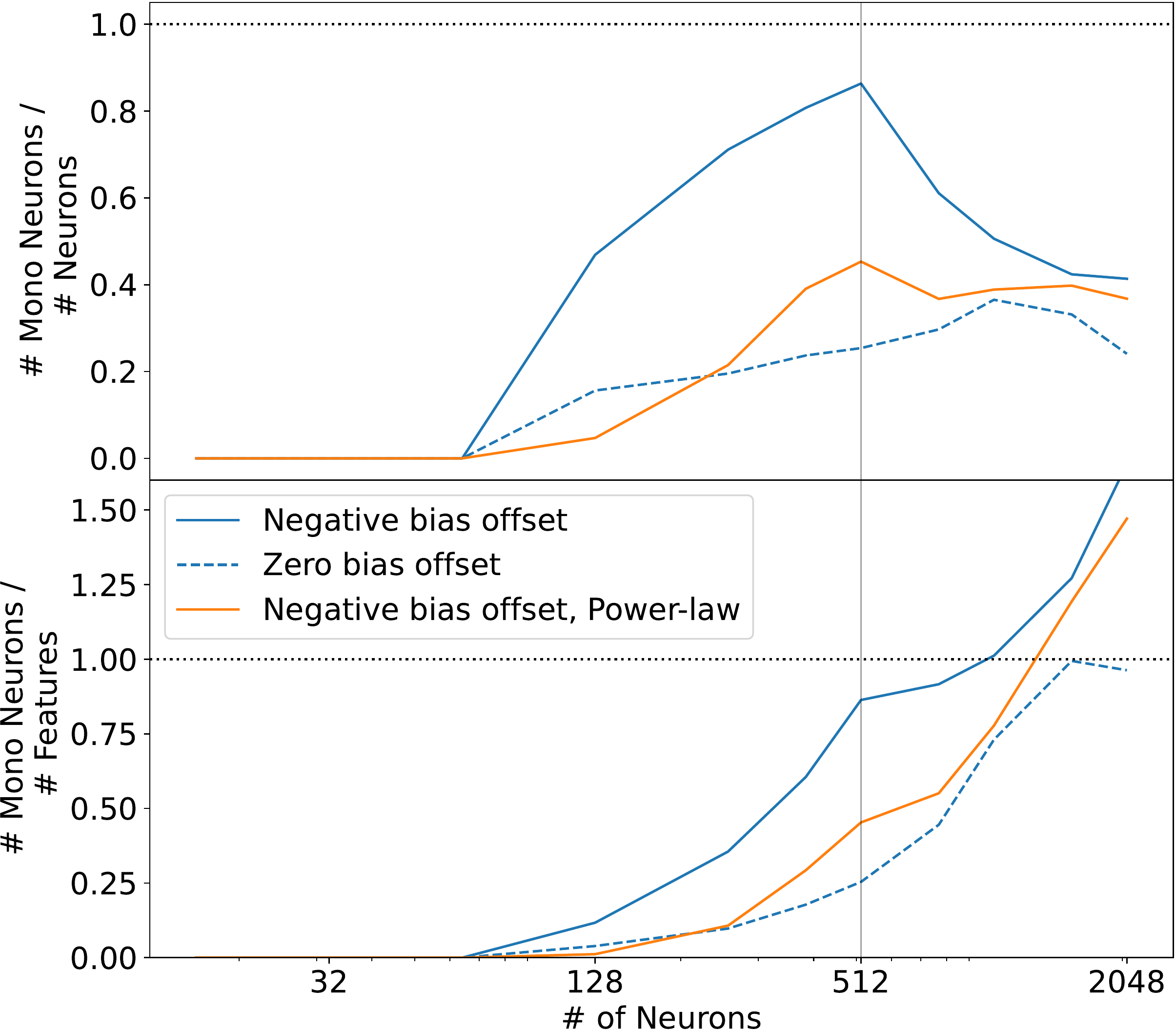}
\caption{The fraction of neurons which are monosemantic (upper) and the ratio of monosemantic neurons to input features (lower) are shown for models with varying numbers of neurons from batches K0 with zero initial mean bias (blue, dashed), K1  with a negative initial mean bias (blue, solid), and K2 with a negative initial mean bias and power-law feature frequencies (orange). The dashed black lines show where the number of monosemantic neurons equals the number of neurons (upper) and the number of features (lower). The vertical line shows where the number of neurons equals the number of input features. The negative initial mean bias indeed increases monosemanticity. We also see that models have more monosemantic neurons as the total number of neurons increases, but that the fraction of neurons which are monosemantic peaks when there is one neuron per feature. We suspect that the fraction of monosemantic neurons would rise in the largest models if we allowed more training time, however, and so this trend may partially reflect the limited training time we allowed.}
\label{fig:k_equal}
\end{figure}

Figure~\ref{fig:k_equal} shows the fraction of neurons which are monosemantic (upper) and the ratio of monosemantic neurons to input features (lower) for models with varying numbers of neurons from batch K0, with zero initial mean bias, batch K1, with a negative initial mean bias, and batch K2, with a negative initial mean bias and power-law feature frequencies.
The vertical line shows where the number of neurons equals the number of input features.
The negative initial mean bias indeed increases monosemanticity.
We see that models have more monosemantic neurons as the total number of neurons increases, supporting claim~\ref{claim:nonlinear_width}.
By contrast, the fraction of neurons which are monosemantic peaks when there is one neuron per feature.
We suspect that the fraction of monosemantic neurons would rise in the largest models if we allowed more training time, however, and so this trend may partially reflect the limited training time we allowed.

Interestingly, a large fraction of the neurons become monosemantic even before there are more neurons than features.
This is not something we see when we do not initialize to a negative bias, showing that an even stronger claim holds: at least for this feature decoding task we can engineer models to be more monosemantic at no cost to the loss even when there are fewer neurons than features.

Given that wider layers allow us to achieve more monosemantic neurons, a natural question is whether we can have wider layers without higher computational costs.
We attempted this by making the linear layers in the model sparse, with fixed connectivity according to an Erd\Horig{o}s-R{\'e}nyi graph.
Unfortunately even modest amounts of sparsity resulted in significantly greater loss, such that dense layers were more performant than sparse ones at fixed compute cost (claim~\ref{claim:sparse_neurons}).
It is possible that more sophisticated approaches could achieve sparsity without such tradeoffs, and we are excited to explore this further in the future.

\subsubsection{Feature Sparsity}\label{sec:feature_sparsity}

\begin{figure}
\centering
\includegraphics[width=\textwidth]{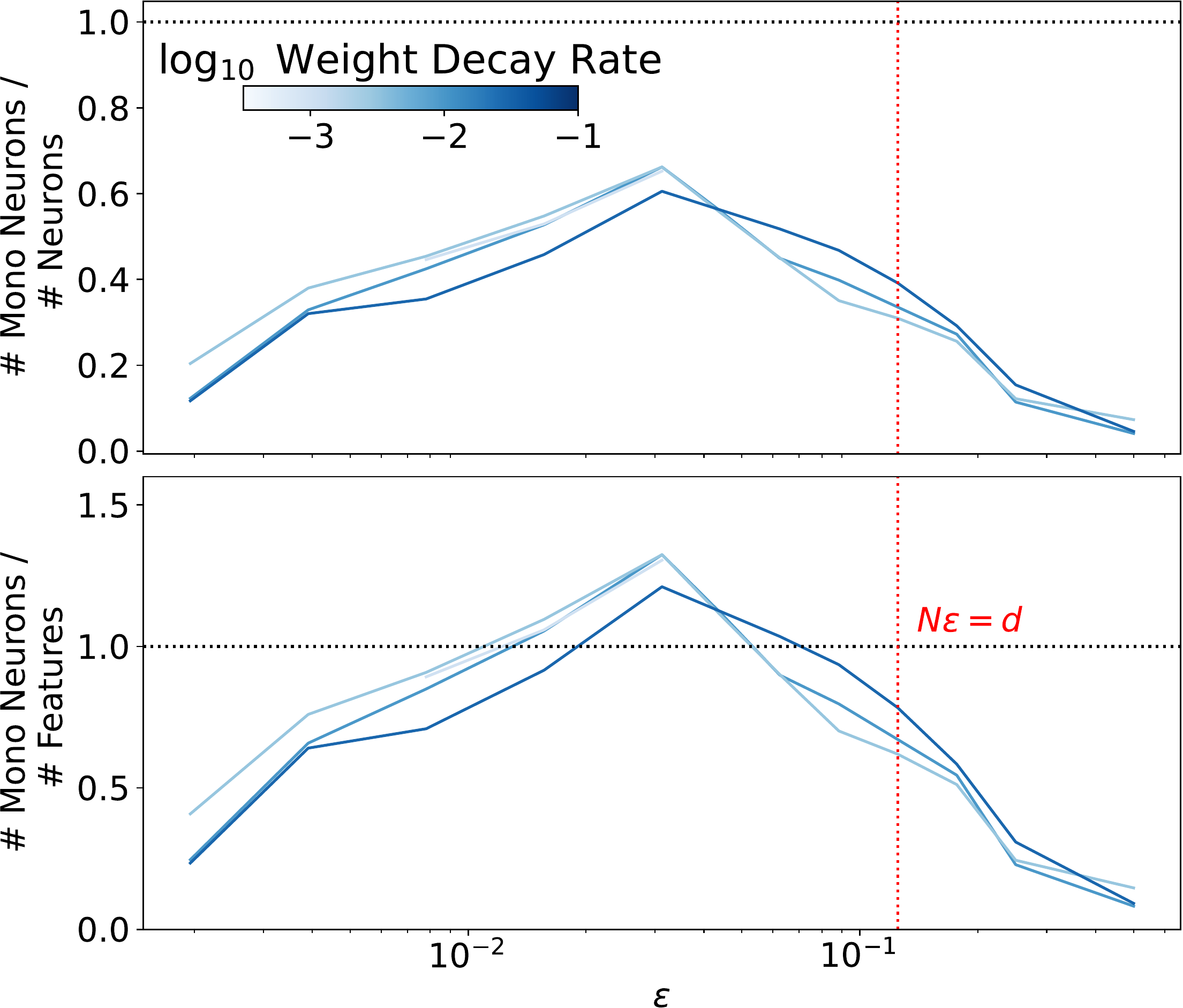}
\caption{The fraction of neurons which are monosemantic (upper) and the ratio of monosemantic neurons to input features (lower) are shown for models from batches E1-E4 as functions of the feature frequency $\epsilon$ and weight decay rate. The dashed black lines show where the number of monosemantic neurons equals the number of neurons (upper) and the number of features (lower). The dashed red line shows where the expected number of features present in a sample ($N\epsilon$) equals the embedding dimension $d$. This line is near the transition from most features being monosemantically represented to most features being polysemantically represented. In the very sparse limit there are highly monosemantic models, but we need to use a slower weight decay and longer training runs to find them because the model sees each feature less often and so learns more slowly.}
\label{fig:eps_sweep_mono}
\end{figure}

So far we have studied models trained in the limit $N\epsilon \ll d$, so that samples are sparse in features.
Recall that $N$ is the number of possible features, $\epsilon$ is the feature frequency, and $d$ is the embedding dimension.
We now examine how models change as we increase the feature frequency $\epsilon$, up through the limit where $N \epsilon \gg d$.

Figure~\ref{fig:eps_sweep_mono} shows the fraction of neurons which are monosemantic (left) and the ratio of monosemantic neurons to input features (right) for these models as functions of the feature frequency.
The red dashed line shows where the expected number of features present in a sample equals the embedding dimension.

\begin{figure}
\centering
\includegraphics[width=\textwidth]{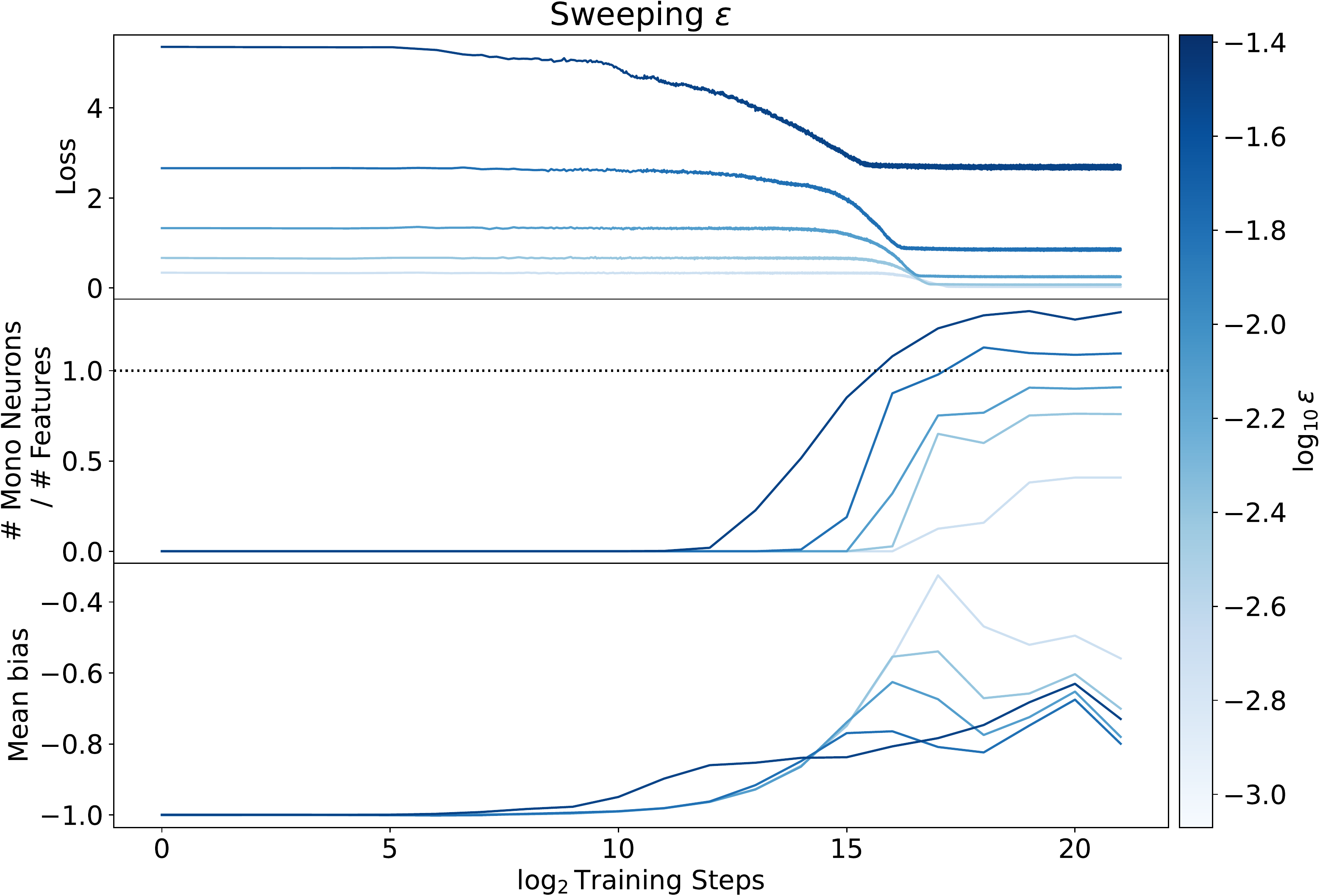}
\caption{Training traces are shown for models with different feature sparsities from batch E3. At higher sparsity we see that the initial mean bias decays further before the loss starts to fall, resulting in models finding less monosemantic local minima.}
\label{fig:eps_training}
\end{figure}

We see a peak in monosemanticity around feature frequency $\epsilon = 2^{-5} \approx 0.03$.
As $\epsilon \rightarrow 0$ training slows, and so a fixed weight decay rate results in the initial negative bias mostly disappearing before the model has a chance to learn most features.
This is illustrated in Figure~\ref{fig:eps_training}, where we see that at higher sparsity the initial mean bias decays further before the loss starts to fall, with the expected result that models find less monosemantic local minima.
This explains the drop in monosemanticity in the $\epsilon \rightarrow 0$ limit.
With slower weight decay (Figure~\ref{fig:eps_sweep_mono}, lighter colors) we see more monosemanticity in this very sparse regime, and we expect that by continuing to scale the weight decay down and training for longer we can achieve uniformly high fractions of monosemantic neurons in this regime, albeit with longer training runs.

As the feature frequency increases (and sparsity falls) models become more polysemantic.
This transition begins around $\epsilon = 2^{-5} $, at which point $N\epsilon = d/4$.
By the time $N\epsilon \approx 2d$ there are almost no monosemantic neurons left. 

We can understand the change in behavior around $N\epsilon = d$ through the lens of compressed sensing.
In general, one can exactly recover an $N$-dimensional $N\epsilon$-sparse vector from a $d$-dimensional projection only if $d = \Omega(N\epsilon \log(1/\epsilon))$ (see the discussion of compressed sensing and references in~\cite{XYZ}).
Because in our case $\log(1/\epsilon)$ is order-unity, this suggests that the kind of recovery that is possible changes around $N\epsilon \approx d$, from exact ($N\epsilon \ll d$) to approximate ($N\epsilon > d$).
It is not too surprising that the model discovers different algorithms in these limits.

\begin{figure}
\begin{adjustwidth}{-2.5cm}{-1.5cm}
\centering
\includegraphics[width=\linewidth]{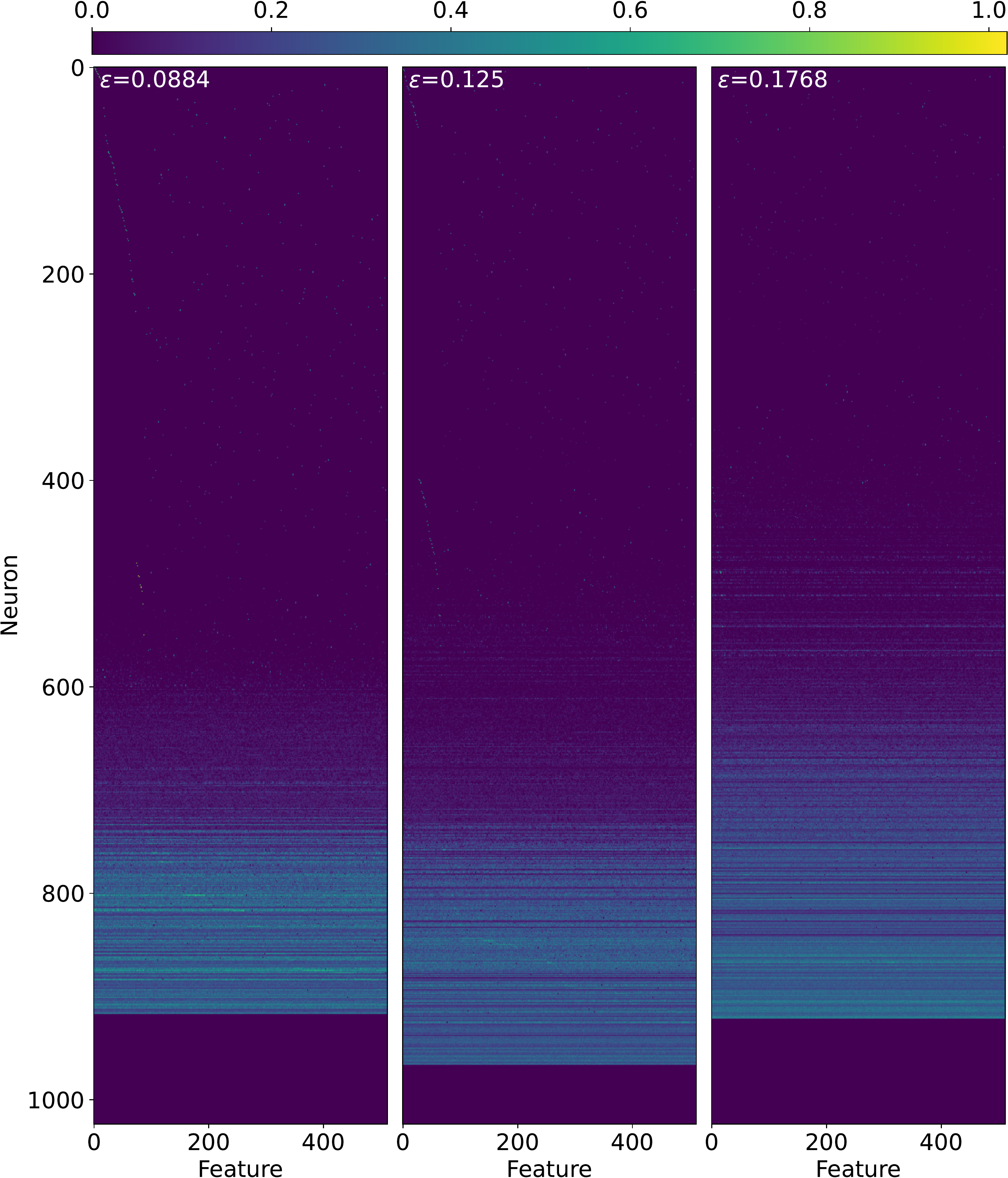}
\end{adjustwidth}
\caption{The neuron activations are shown for all single-feature inputs of unit strength. Neurons are sorted by monosemanticity, with the upper-most being the most monosemantic (equation~\ref{eq:r}).  Features are sorted by the neuron they activate most-strongly. The different panels show this for different final models from batch E1 with different feature frequencies. No sharp phase transition is evident: rather models appear to smoothly transition from highly-monosemantic to highly-polysemantic across the boundary between the feature-sparse and feature-dense limits.}
\label{fig:eps_sfa_plot}
\end{figure}

We have inspected the single-feature activations of models near the boundary $N\epsilon=d$ (Figure~\ref{fig:eps_sfa_plot}).
We see a smooth rise in the number and activation magnitude of heavily-polysemantic neurons and a steady fall in the number and activation magnitude of monosemantic neurons.
There is no sign of a sharp phase transition: it appears that models smoothly transition from a highly monosemantic algorithm to a highly polysemantic one.

\subsection{Task: Random Re-Projector}\label{sec:reproj}

\begin{figure}
\centering
\includegraphics[width=\textwidth]{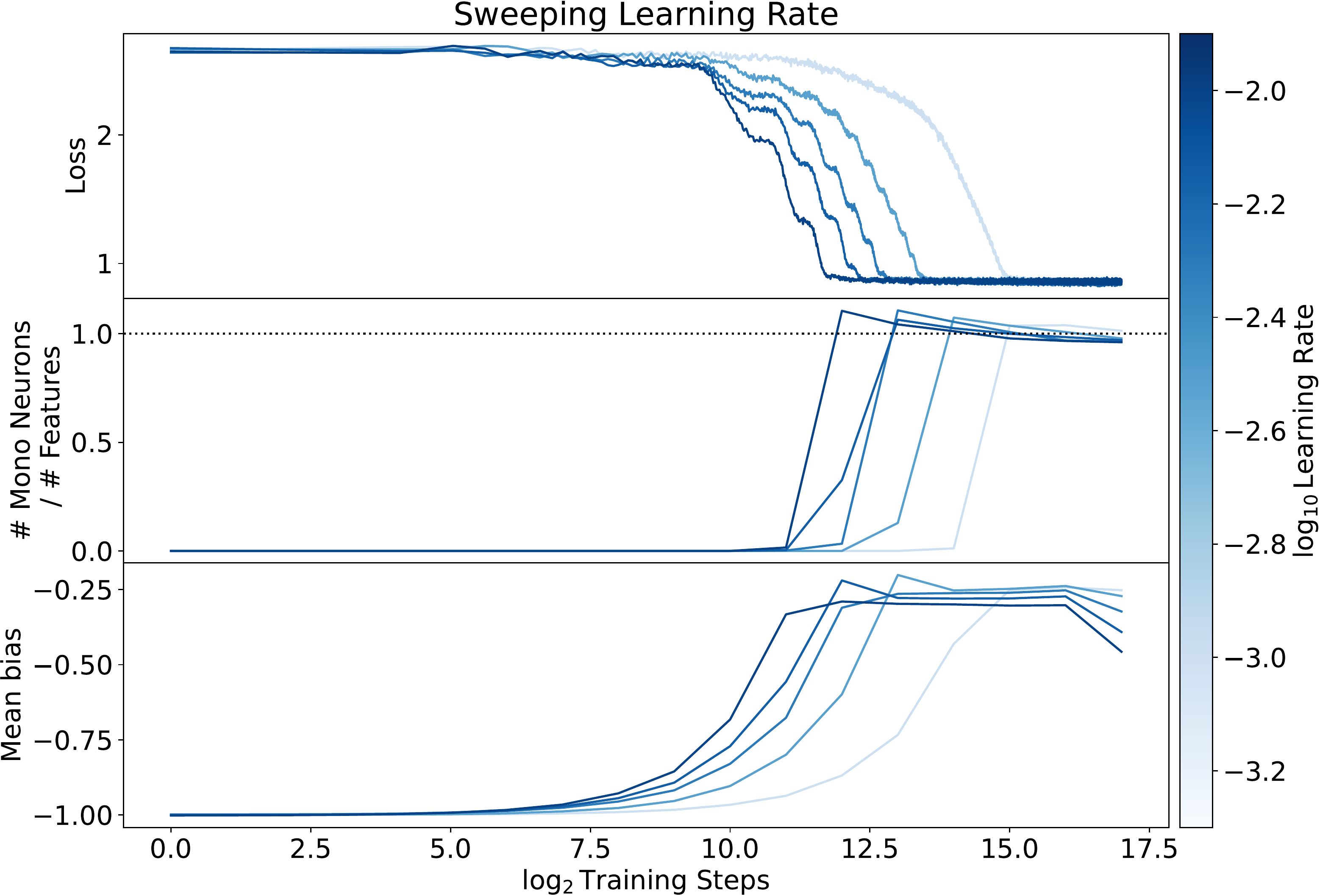}
\caption{Training traces are shown for random re-projector models with different learning rates with from batch RP1. As in the decoder models, with an initial negative bias we are able to create models with roughly one monosemantic neuron per feature across a range of learning rates.}
\label{fig:reproj_lr}
\end{figure}

For the random re-projector task we have similar findings.
Once more, with the negative initial mean bias we are able to create models with roughly one monosemantic neuron per feature across a range of learning rates (Figure~\ref{fig:reproj_lr}).

\subsection{Task: Absolute Value}\label{sec:abs}

\begin{figure}
\centering
\includegraphics[width=\textwidth]{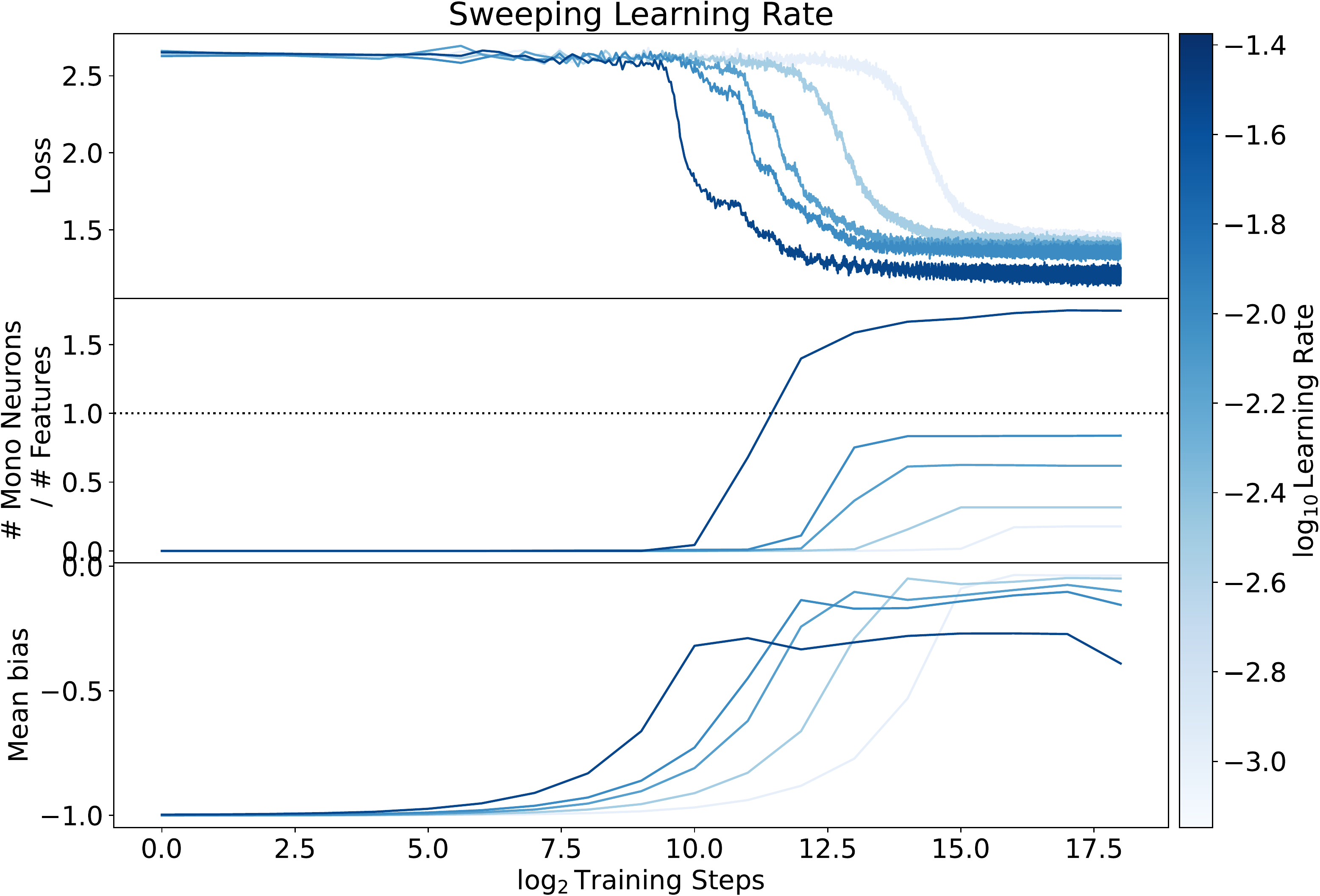}
\caption{Training traces are shown for absolute value models with different learning rates with from batch LR5. Unlike in the decoder and re-projector models, we see that basins of different monosemanticity have different losses, with lower losses for more-monosemantic solutions. We also see here that a negative initial bias alone is not enough to ensure a highly monosemantic solution: the monosemanticity rises with increasing learning rate.}
\label{fig:abs_lr}
\end{figure}

\begin{figure}
\centering
\includegraphics[width=\textwidth]{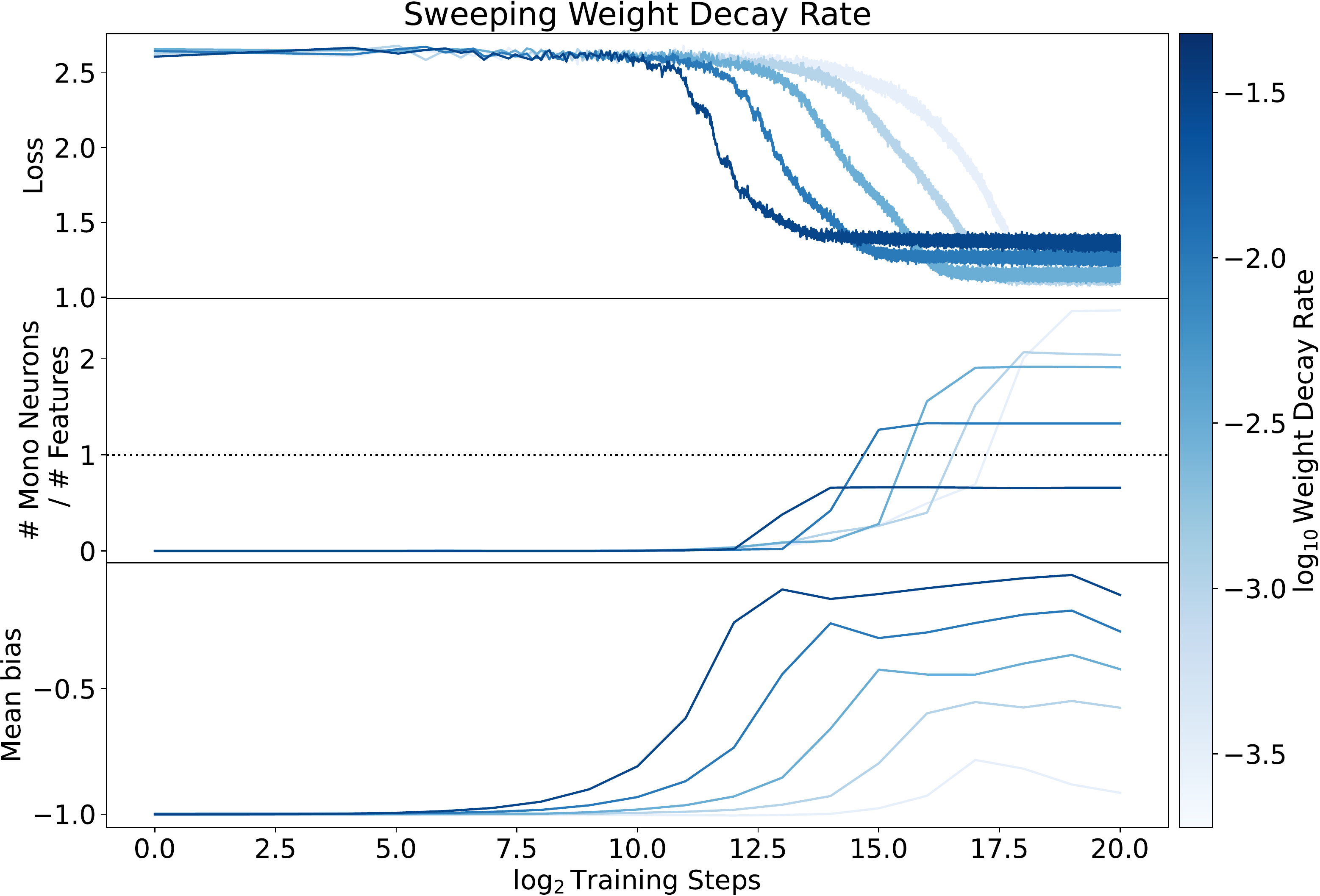}
\caption{Training traces are shown for absolute value models with different weight decay rates with from batch D1. Slower weight decay rates produce more monosemantic solutions.}
\label{fig:abs_decay}
\end{figure}

For the absolute value task we have similar findings.
Once more we see that there are both polysemantic and monosemantic basins, and that increasing the learning rate results in increasingly monosemantic models (Figure~\ref{fig:abs_lr}).
We also see that the more monosemantic models have larger negative biases.
There are a few key differences, however:
\begin{enumerate}
	\item Models of different monosemanticity achieve different loss, with lower losses for more monosemantic models.
	\item Setting a negative initial bias is insufficient to robustly produce mostly-monosemantic models. For instance, in Figure~\ref{fig:abs_lr} we used an initial bias of $-1$ and yet several of the models shown are largely polysemantic.
\end{enumerate}

In exploring these differences, we also determined that using less weight decay produced more monosemantic models (Figure~\ref{fig:abs_decay}).
The combination of a negative initial bias and a slow rate of weight decay appears sufficient to make highly monosemantic models.
We hypothesize that the slower weight decay is necessary because this task is more complicated than the decoder task, and so it takes longer to learn.
As a result, with fast weight decay the model passes too quickly through the region of negative biases where it \emph{could} learn the monosemantic basin.

This, as well as out finding in Section~\ref{sec:feature_sparsity} that slower weight decay leads to more monosemantic models, suggests that there could be value in exploring other approaches, as very slow weight decay rates end up requiring very long training runs to converge to the loss plateau.
One such approach which we are excited about is regularizing towards a negative bias rather than initializing with a negative bias, which may lessen the path-dependence of the training process.
We have run preliminary experiments with such regularizers and find that they indeed produce more monosemantic models than the ones which begin with zero mean bias, though there remain open questions about e.g. how strong a regularizer to use.

\section{Mechanistic Interpretability}\label{sec:interp}

We now turn to the mechanistic interpretation of our models.
Here we focus on the feature decoder, as it is the simpler model to interpret, and we focus specifically on the feature decoders which learn highly-monosemantic representations.

To start, we divide our models into two pieces, one containing all neurons with positive bias and one containing all neurons with negative bias. This approximately divides the models into polysemantic and monosemantic neurons respectively.
With this division we can study different contributions to the model output.

\subsection{Monosemantic Neurons}

\begin{figure}
\centering
\includegraphics[width=\textwidth]{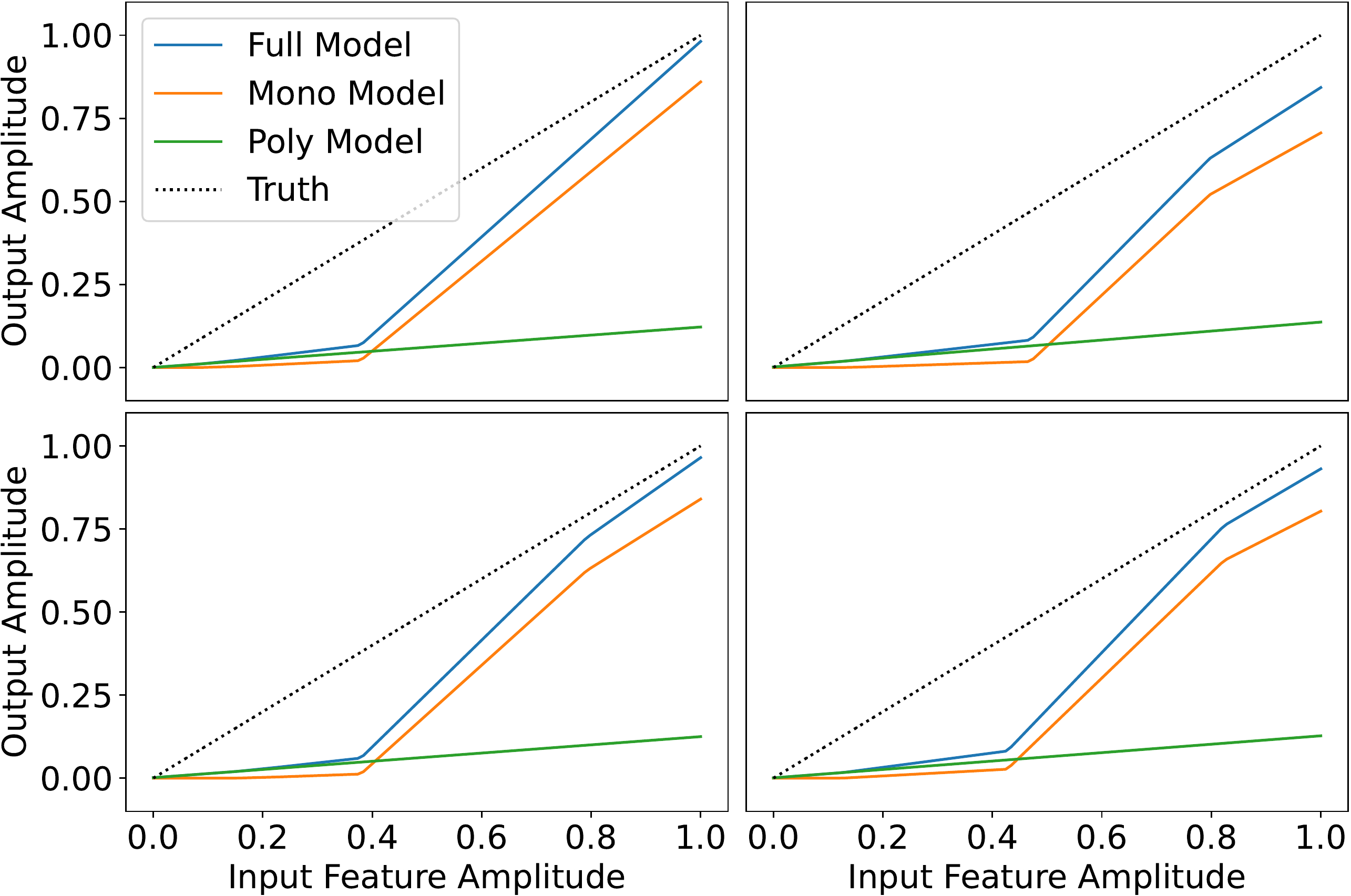}
\caption{The outputs of a final model from batch E1 with $\epsilon=1/64$ are shown for four features as a function of input feature amplitude for samples containing just one active feature. The output is shown for the full model (blue), the monosemantic portion (orange), and the polysemantic portion (green). Exact recovery is marked by the dotted line. The upper left panel shows a feature primarily represented by a single monosemantic neuron, while the remaining three panels show features primarily represented by a pair of monosemantic neurons each. Note that the additional neuron is not there for redundancy, rather it allows the model to express a more fine-grained notion of confidence in the feature.}
\label{fig:amplitude_plot}
\end{figure}

Next, we can study how different features are represented as a function of feature amplitude.
Figure~\ref{fig:amplitude_plot} shows the outputs of a final model from batch E1 with $\epsilon=1/64$.
This is shown for four features as a function of input feature amplitude for samples containing just one active feature.
The output is shown for the full model (blue), the monosemantic portion (orange), and the polysemantic portion (green), and exact recovery is marked by the dotted line.

The upper left panel shows a feature primarily represented by a single monosemantic neuron.
This has a negative bias, and so only activates once the feature amplitude is somewhat greater than zero.
This makes sense: for a model trained with many features active per sample, small apparent feature amplitudes are likely to be due to interference rather than a true input feature, and so it is better to output zero than apparent amplitude.

Above the zero cutoff, the monosemantic part of the output for the upper-left feature rises quickly, approaching the ground truth line.
The output is always below this line because there is always a chance that the feature is not really present, and so it is optimal to output something slightly less than the apparent amplitude.

Note that, because a ReLU only permits one kink in the output per neuron, features which are primarily represented by a single neuron only support one kink.
This gets spent, in effect, on the model correcting the low-amplitude end for the possibility of interference, so the model has no choice but to fit a line for the high-amplitude end.
Such a line necessarily either undershoots or overshoots the optimal output over a wide range, because it begins below the optimum, and so there is room for improvement.

We see an example of such improvement in the other three panels, which show three features each of which is represented primarily by a pair of monosemantic neurons.
With access to two neurons the model can kink the output line twice, resulting in three regions: a region with zero output, a region with rapidly rising output, and a region at high input amplitude where the output amplitude closely tracks the input amplitude.
We can interpret the first region as before: low-amplitude features are likely due to interference and so best ignored.
Above some input amplitude the feature is almost certainly real, so the third region just tracks the ground truth with a small downward correction to allow for the non-zero probability that the feature is not really present.
Finally, the second region interpolates between these two.

\begin{figure}
\centering
\includegraphics[width=\textwidth]{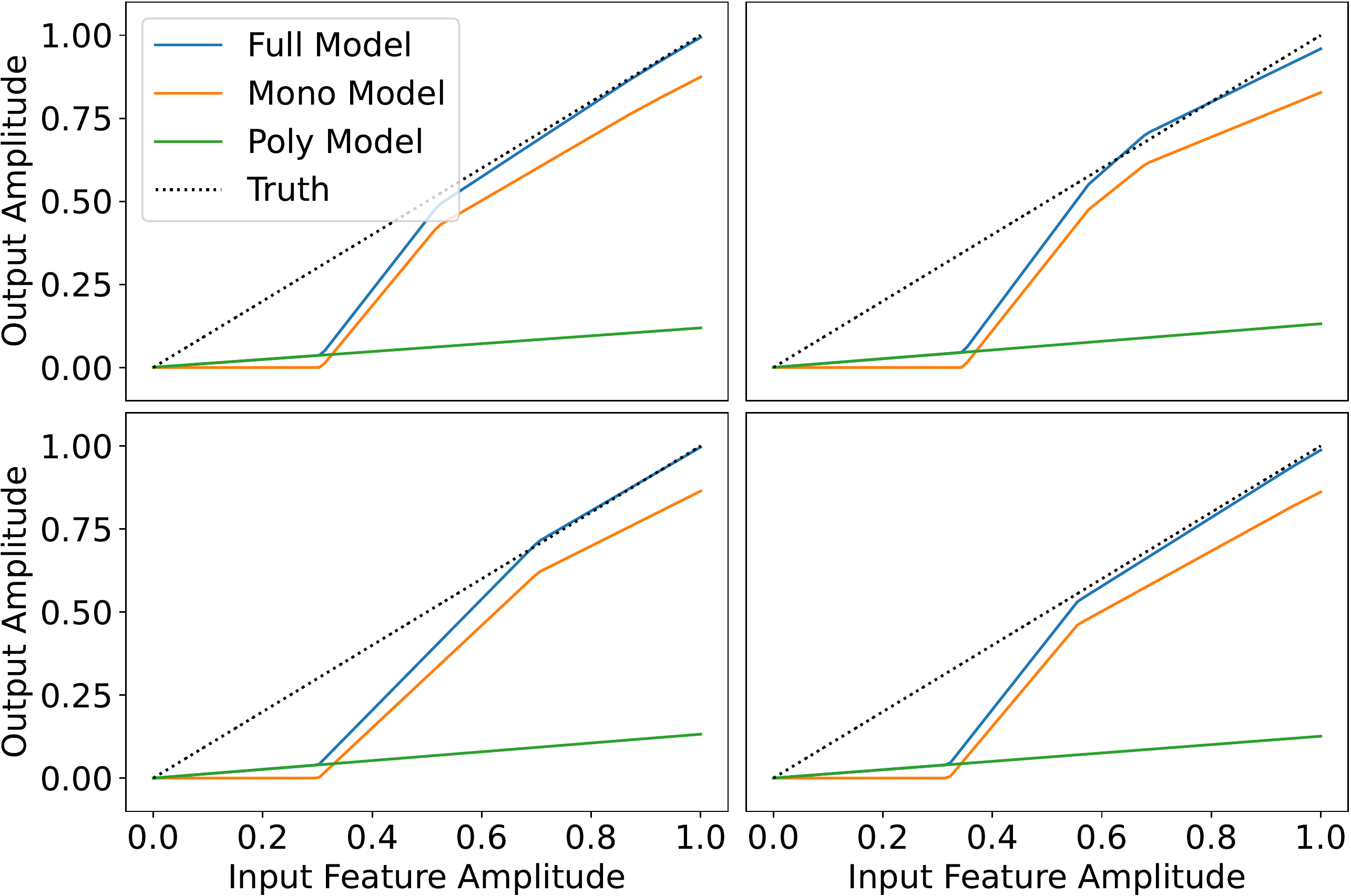}
\caption{The outputs of a final model from batch E3 with $\epsilon=1/256$ are shown for two features as a function of input feature amplitude for samples containing just one active feature. The output is shown for the full model (blue), the monosemantic portion (orange), and the polysemantic portion (green). Exact recovery is marked by the dotted line. Because this model was trained with a lower feature frequency than that of Figure~\ref{fig:amplitude_plot}, it is more confident that the features are not due to interference and so produces non-zero output for lower-amplitude inputs and produces outputs which are closer to the ground truth for strong-amplitude inputs.}
\label{fig:amplitude_plot_sparser}
\end{figure}

Figure~\ref{fig:amplitude_plot_sparser} provides additional evidence for this interpretation.
There we show the same quantities for a model trained at lower feature frequency.
With decreasing feature frequency (increasing sparsity), interference becomes less frequent and so the model comes to favor representing features closer to the ground truth.
We see this in three distinct ways:
\begin{enumerate}
\item The input amplitude at which the monosemantic neurons produce a non-zero output is $\approx 0.3$ versus $\approx 0.4$ for the higher-frequency model in Figure~\ref{fig:amplitude_plot}.
\item For high-amplitude inputs, the model outputs amplitudes closer to the ground truth (input amplitudes) than the higher-frequency model of Figure~\ref{fig:amplitude_plot} does.
\item The features which are represented primarily by two monosemantic neurons have their second kink at lower input amplitudes than those in the higher-frequency model of Figure~\ref{fig:amplitude_plot}, down from $\approx 0.8$ to $\approx 0.6$.
\end{enumerate}
All three of these differences reflect an increased confidence that features are as they appear and are not due to interference.

\subsection{Polysemantic Neurons}\label{sec:poly}

With a theory in hand for the monosemantic neurons, we now turn to the polysemantic neurons.
In our plots above these always appear as solid straight lines.
Inspecting the model from Figure~\ref{fig:amplitude_plot} in more detail, we find that this is not far off.
For most features, on a single-feature input the polysemantic neurons collectively produce a contribution which begins very near zero ($< 0.005$), rises with a modest nearly-constant slope ($0.09-0.16$, varying typically $<0.01$ over the amplitude interval $[0,1]$).

As far as we can tell, the purpose of the polysemantic neurons is primarily to ``tilt'' the output of each feature up slightly, so that e.g. at low input amplitude the output is slightly non-zero.
This correction helps the model to better reflect its confidence in each feature being real: because features have input amplitudes ranging from 0-1, the optimal output is never zero even if the feature is very unlikely to be present.
This also explains the modest ranges of slopes and intercepts for the polysemantic contribution across features: if the polysemantic neurons are accounting for the same adjustment in confidence on every feature it makes sense for them to produce the same or very similar corrections to the corresponding outputs.

\begin{figure}
\centering
\includegraphics[width=\textwidth]{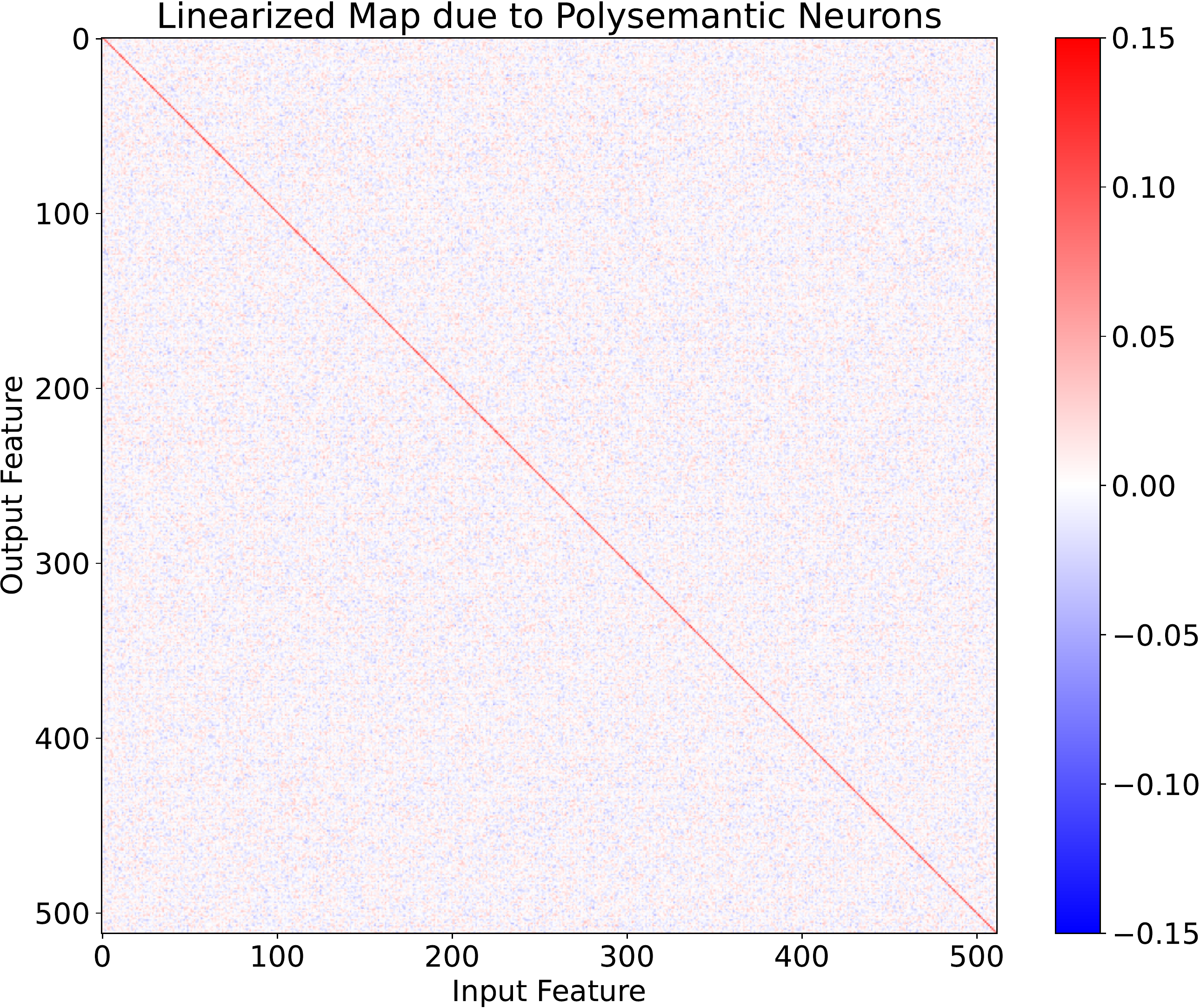}
\caption{The linearized map from input features to output features is shown for the polysemantic (positive-bias) neurons of a final model from batch E1 with $\epsilon=1/64$. This is close to a representation of the identity, with singular values that are all either zero or nearly-unity ($0.99-1.03$). This provides evidence that the polysemantic neurons implement a linear model with similar slopes across neurons.}
\label{fig:poly_linear}
\end{figure}

Further evidence for this interpretation is provided by the linearized map implemented by the polysemantic neurons, shown in Figure~\ref{fig:poly_linear}.
This was computed by multipling the fixed random projection matrix with the first-layer weights leading into positive-bias neurons, and multiplying that by the second-layer weights leading out of positive-bias neurons.
The result is very nearly a low-rank representation of the identity matrix, with singular values which are all either zero or near-unity ($0.99-1.03$).
This is what we would expect if the polysemantic neurons have learned to implement a linear model with similar slopes for each feature.
This also explains why the number of polysemantic neurons is very nearly the embedding dimension: the rank of the representation of the identity is bounded by the embedding dimension, so there are no gains to be had by allocating more neurons to this representation.

Of course the polysemantic neurons do not implement a purely linear model.
In particular, the product of the fixed projection matrix and the first-layer weights can produce vectors with negative entries, and those will be set to zero by the ReLU before passing into the final layer.
This explains why the polysemantic neurons have modest positive biases: this avoids the model from running into the negative realm of the ReLU.
These biases do not contribute much on their own to the output because the model has contrived for the inner product of the positive biases with the second-layer weights to be near zero ($< 0.005$ in magnitude).

\subsection{Summary}

In summary, when our feature decoder models learn highly monosemantic representations, they implement the following algorithm:
\begin{itemize}
	\item Monosemantic neurons are assigned to features.
	\item When a feature is represented by a single monosemantic neuron, that neuron de-noises interference for small inputs and otherwise outputs an estimate of the input feature amplitude (weighted by the probability of spurious features due to interference).
	\item When a feature is represented by two monosemantic neurons, the extra degree of freedom is used to make the estimate of the input feature amplitude more accurate at the high-amplitude end.
	\item The polysemantic neurons implement a linear model for the input amplitudes on the training distribution. This model is a low-rank representation of the identity, which increases each feature's output slightly in proportion to its input strength, producing a better approximation of the optimal estimator.
\end{itemize}
We have not performed an equivalent analysis on the highly-polysemantic models nor on the other tasks.
We expect that a similar kind of analysis would yield insights into the absolute value task, but are less hopeful for the highly-polysemantic models.

\section{Conclusions}\label{sec:conclusions}

We have found that when inputs are feature-sparse, models can be made more monosemantic without increasing the loss by just changing which loss minimum the training process finds.
More monosemantic loss minima have moderate negative biases in the tasks we studied, and we were able to use this fact to engineer highly monosemantic models.
We expect this property to be quite generic among feature decoding/identification tasks, where the negative bias serves as a guard against interference from other features.
As such, we expect this approach of using negative biases to apply to a broader set of models than the toy ones we studied here.

Unfortunately, our current engineering approach is somewhat fragile because it depends on setting a bias weight decay rate which must be slow relative to the rate at which a model learns a task.
We believe that this can be remedied by tailoring the bias decay on a per-neuron basis, or tying the bias decay to the rate of change of the rest of the model weights, or by using a regularizer towards negative bias instead of relying on a negative initial mean bias. 
Our sense is that there is much low-hanging fruit here, and we intend to explore this further in future work.

We further found that providing models with more neurons per layer makes them more monosemantic, albeit at increased computational cost.
We hoped that naively introducing sparsity in the model weights would allow us to reduce the cost to be competitive with narrower, dense models while achieving comparable task perforamnce.
Instead, we found that sparsity resulted in significant reductions in task performance at comparable compute cost.
Thus, if we are to use this finding that wider nonlinear layers are more monosemantic, we need some other approach to lowers the compute cost per neuron without degrading performance excessively.
This is a key direction for future work.

Finally, we still do not understand the algorithm implemented by our most polysemantic models.
The neurons in these models are very polysemantic typically firing for \emph{most} features (e.g. Figure~\ref{fig:ReLU_equal_lr_sfa_zero_bias}, left), and such neurons make up a large fraction of the total.
It is possible that there is a way to understand what these neurons are doing in this toy model setting where we know the ground truth of what the features are.
If we are able to do this, there may be insights we can transfer to more real-world settings, and so the final key direction we have identified is to attempt to unravel the computations implemented by these highly polysemantic neurons.

\section*{Author Contributions}

\begin{itemize}

\item \textbf{Basic Results:} The basic results and experiments demonstrating the existence of monosemantic and polysemantic minima and the path-dependence of the training process were done by ASJ.

\item \textbf{Toy Model:} The toy model was designed by ASJ.

\item \textbf{Training:} The training process were designed by ASJ with input from NS. NS contributed suggestions on the Lamb optimizer, the learning rate schedule, and general code optimizations.

\item \textbf{Bias:} The identification of bias as a signal of monosemanticity was done jointly by ASJ and NS. The use of this to engineer more monosemantic models was devised and implemented by ASJ, with input on the weight decay procedure from NS.

\item \textbf{Activation Functions:} The investigation of different activation functions was done by ASJ with guidance and insights from NS and EH.

\item \textbf{Sparsity and Layer Width:} The discovery that increasing layer width and increasing feature sparsity both increase monosemanticity was done by ASJ.

\item \textbf{Mechanistic Interpretation:} The interpretation of the trained models was done by ASJ with feedback from NS and EH. EH suggested the approach of separately characterizing the input/output maps of the polysemantic and monosemantic neurons.

\item \textbf{Writing and Illustration:} The paper was written and illustrated primarily by ASJ, with suggestions from NS and EH.

\end{itemize}

\section*{Acknowledgments}
We are grateful to Chris Olah for encouragement as well as suggestions including the random re-projector task, the idea to sort features by which neuron they most activate, and the idea to compare with L1 regularization.
Thanks to Sam Marks, and Xander Davies for discussions on this project.
Thanks to Buck Shlegeris, Adam Scherlis, and Kshitij Sachan for discussions on polysemanticity.
Thanks to Nova DasSarma, Shauna Kravec, and Andrei P{\"o}hlmann of Hofvarpnir Studios for computation resources and support in this project.

\appendix

\section{Training Details}\label{sec:training}

Training was performed using the LAMB optimizer~\cite{DBLP:journals/corr/abs-1904-00962} to take advantage of larger batch sizes.
The batch size was chosen empirically to saturate GPU usage, giving size $b = 2^{23}/k$, where $k$ is the size of the nonlinear layer.

The learning rate was controlled with a cosine annealing schedule with $T_{\rm min}=0$ and $T_{\rm max}=2^9$.
We found that this choice converged slightly faster than other choices like a constant learning rate.

An implementation of our model and training process is available on~\href{https://github.com/adamjermyn/toy_model_interpretability}{GitHub}.
Table~\ref{table:batch} provides a full list of the training parameters and model architectures we used in these experiments.

\begin{table}
\caption{Training parameters and model architectures.\label{table:batch}}
\begin{adjustwidth}{-1.5in}{-1.5in}
\begin{tabular}{cccccccccccc}
\hline
\hline\vspace{-0.3em}\\
Batch & Task & Activation & Feature Dist. & $N$ & $m$ & $k$ & $\epsilon$ & Learning Rate & Decay Rate & Bias Offset & L1 Reg.\\
\hline
 LR1 & Decoder & ReLU & Uniform & 512 & 64 & 1024 & 1/64 & Variable & 0 & 0 & 0\\
 LR2 & Decoder & ReLU & Power-law & 512 & 64 & 1024 & 1/64 & Variable & 0 & 0& 0\\
 LR3 & Decoder & ReLU & Uniform & 512 & 64 & 1024 & 1/64 & Variable & 0.03 & -1& 0\\
 B1 & Decoder & ReLU & Uniform & 512 & 64 & 1024 & 1/16 & 0.003 & 0.03 & Variable& 0\\
 B2 & Decoder & ReLU & Uniform & 512 & 64 & 1024 & 1/32 & 0.003 & 0.003 & Variable& 0\\
 B3 & Decoder & ReLU & Uniform & 512 & 64 & 1024 & 1/64 & 0.003 & 0.003 & Variable& 0\\
 B4 & Decoder & ReLU & Uniform & 512 & 64 & 1024 & 1/128 & 0.003 & 0.003 & Variable& 0\\
 B5 & Decoder & ReLU & Uniform & 512 & 64 & 1024 & 1/256 & 0.003 & 0.003 & Variable& 0\\
 LR4 & Decoder & ReLU & Power-law & 512 & 64 & 1024 & 1/64 & Variable & 0.03 & -1& 0\\
 B3 & Decoder & GeLU & Uniform & 512 & 64 & 1024 & 1/64 & Variable & 0.03 & Variable& 0\\
 E1 & Decoder & ReLU & Uniform & 512 & 64 & 1024 & Variable & 0.003 & 0.03 & -1& 0\\
 E2 & Decoder & ReLU & Uniform & 512 & 64 & 1024 & Variable & 0.003 & 0.01 & -1& 0\\
 E3 & Decoder & ReLU & Uniform & 512 & 64 & 1024 & Variable & 0.003 & 0.003 & -1& 0\\
 E4 & Decoder & ReLU & Uniform & 512 & 64 & 1024 & Variable & 0.003 & 0.001 & -1& 0\\
 K0 & Decoder & ReLU & Uniform & 512 & 64 & Variable & 1/64 & 0.007 & 0 & 0& 0\\
 K1 & Decoder & ReLU & Uniform & 512 & 64 & Variable & 1/64 & 0.007 & 0.03 & -1& 0\\
 K2 & Decoder & ReLU & Power-law & 512 & 64 & Variable & 1/64 & 0.007 & 0.03 & -1& 0\\
 RG1 & Decoder & ReLU & Uniform & 512 & 64 & 1/64 & 0.005 & 0.03 & -1& Variable\\
 RP1 & Re-Projector & ReLU & Uniform & 512 & 64 & 1/64 & Variable & 0.03 & -1& 0\\
 LR5 & Abs. & ReLU & Uniform & 512 & 64 & 2048 & 1/64 & Variable & 0.03 & -1& 0\\
 D1 & Abs. & ReLU & Uniform & 512 & 64 & 2048 & 1/64 & 0.007 & Variable & -1& 0\\
\hline
\end{tabular}
\end{adjustwidth}
\end{table}

\clearpage

\bibliographystyle{halpha}
\bibliography{main}

\end{document}